\def\isarxiv{1}

\ifdefined\isarxiv

\documentclass[11pt]{article}

\usepackage[top=1in, bottom=1in, left=0.75in, right=0.75in]{geometry}

\DeclareSymbolFont{rsfs}{U}{rsfs}{m}{n}
\DeclareSymbolFontAlphabet{\mathscrsfs}{rsfs}
\usepackage{hyperref}
\usepackage{natbib}
\else
\documentclass{article}
\usepackage{multicol,enumitem}

\fi
\usepackage[noend]{algpseudocode} %
\usepackage{algorithm} %

\usepackage{url}   

\usepackage{amsmath}
\usepackage{amssymb}
\usepackage{amsthm}

\usepackage{booktabs}       %
\usepackage{amsfonts}       %

\usepackage{dsfont,mathtools,bm}
\usepackage{nicefrac}       %
\usepackage{microtype}      %
\usepackage{xcolor}         %

\usepackage{amsmath,amsthm,amssymb}
\usepackage{cleveref}
\usepackage{comment}
\usepackage{bbm}
\usepackage{textcomp}
\usepackage{wrapfig}
\usepackage{float}
\usepackage{subcaption}
\hypersetup{
  colorlinks   = true, %
  urlcolor     = blue, %
  linkcolor    = blue, %
  citecolor   = blue %
}

\usepackage{url}            %
\usepackage{booktabs}       %
\usepackage{amsfonts}       %
\usepackage{nicefrac}       %
\usepackage{microtype}      %
\usepackage{multirow}
\usepackage{amsmath}
\usepackage{amsthm}
\usepackage{amssymb}

\usepackage{color}
\usepackage[english]{babel}
\usepackage{graphicx}
\usepackage{grffile}
\usepackage{wrapfig,epsfig}
\usepackage{url}
\usepackage{color}
\usepackage{algpseudocode}
\usepackage[T1]{fontenc}
\usepackage{bbm}
\usepackage{comment}
\usepackage{tikz}

\usepackage{pgfplots}
\pgfplotsset{compat=1.8}
\tikzset{elegant/.style={smooth,thick,samples=500,magenta}}

\theoremstyle{plain}
\newtheorem{theorem}{Theorem}[section]

\theoremstyle{definition}

\newtheorem{assumption}[theorem]{Assumption}

\crefname{assumption}{Assumption}{Assumptions}

\theoremstyle{plain}
\newtheorem*{thm*}{Theorem}

\theoremstyle{plain}

\newcommand{\M}{\mathcal{M}}

\newcommand{\V}{\mathcal{V}}

\newcommand{\mask}{\mathrm{[M]}}
\newcommand{\info}{\mathrm{info}}

\newcommand{\name}{\ifmmode\mathrm{ETE}\else ETE\fi}

\newcommand{\fullname}{\ifmmode\textbf{E}\text{xplore-}\textbf{T}\text{hen-}\textbf{E}\text{xploit}\else \textbf{E}xplore-\textbf{T}hen-\textbf{E}xploit\fi}

\definecolor{b2}{RGB}{51,153,255}
\definecolor{myGreen}{RGB}{80,180,0}
\definecolor{myGold}{rgb}{0.75,0.6,0.12}

\newcommand{\bx}{\mathbf{x}}

\title{From Bits to Rounds: Parallel Decoding with Exploration for Diffusion Language Models}

\ifdefined\isarxiv
\author{Hengyu Fu\thanks{University of California, Berkeley. Email: \texttt{\{hengyuf,baihe\_huang,jiantao\}@berkeley.edu}} \footnotemark[2] \;\; Baihe Huang\footnotemark[1] \thanks{Equal Contributions.}  \;\;  Virginia Adams \thanks{NVIDIA. Email: \texttt{\{jiantaoj,vadams,charlwang,venkats\}@nvidia.com}}  \\  \\ Charles Wang\footnotemark[3]  \;\; 
Venkat Srinivasan\footnotemark[3] \;\; Jiantao Jiao\footnotemark[1] \footnotemark[3] }
\date{}

\else

\author{%
  David S.~Hippocampus\thanks{Use footnote for providing further information
    about author (webpage, alternative address)---\emph{not} for acknowledging
    funding agencies.} \\
  Department of Computer Science\\
  Cranberry-Lemon University\\
  Pittsburgh, PA 15213 \\
  \texttt{hippo@cs.cranberry-lemon.edu} \\
}

\fi

\begin{document}

\setlength{\abovedisplayskip}{3.2pt}
\setlength{\belowdisplayskip}{3.2pt}

\ifdefined\isarxiv
  \maketitle

\begin{abstract}

Diffusion Language Models (DLMs) have recently emerged as a strong alternative to autoregressive language models (LMs). DLMs offer comparable accuracy with faster inference speed via parallel decoding.
However, standard DLM decoding strategies relying on high-confidence tokens encounter an inherent information-theoretic bottleneck that restricts decoding progress and ultimately slows generation.
We demonstrate both theoretically and empirically that prioritizing high-confidence tokens is inherently inefficient. High-probability tokens carry negligible information and strictly relying on them limits the effective progress made in each decoding round.
We prove that the number of decoding rounds must grow linearly with the sample's total information (negative log-likelihood) and inversely with the per-round information budget, establishing a bits-to-rounds principle.
We also propose Explore-Then-Exploit (ETE), a training-free decoding strategy that maximizes information throughput and decoding efficiency. ETE combines cross-block decoding with targeted exploration of high-uncertainty tokens to reshape the conditional distribution and trigger cascades of confident predictions.
Experiments verify our theoretical bounds and demonstrate that ETE consistently reduces the required number of decoding rounds compared to confidence-only baselines without compromising generation quality.

\end{abstract}
\tableofcontents

\else
\maketitle
\begin{abstract}

\end{abstract}

\fi

\section{Introduction}

Diffusion Language Models (DLMs) exploit bidirectional attention to decode multiple tokens simultaneously (i.e., parallel decoding). At the start of generation, all output tokens are masked. In each decoding round, DLMs iteratively unmask a set of potentially non-sequential tokens until the entire output sequence has been uncovered. This ability to generate multiple tokens per forward pass yields substantial throughput gains compared to the inherently sequential decoding process of standard autoregressive models. Many popular diffusion models employ a hybrid decoding strategy in which the output sequence is divided into equal-sized blocks, and each block undergoes the diffusion unmasking process sequentially. Prior works have found this block-based decoding approach to deliver the best flexibility between speed and accuracy when compared with full parallel decoding and autoregressive decoding.

Empowered by Masked Diffusion Model (MDM) architectures,
recent works~\cite{nie2025large,zhu2025llada,prabhudesai2025diffusion} have demonstrated strong scalability and competitive performance of DLMs across a broad range of tasks. Moreover, commercial DLMs~\citep{google2025geminidiffusion, khanna2025mercury,song2025seed} have made significant industrial impact through superior inference throughput and speed-quality tradeoffs.

Despite these successes, DLM inference faces a fundamental challenge: parallel decoding incurs inherent performance degradation. Independently sampling tokens from diffusion LM marginal distributions does not reproduce the true joint distribution~\cite{feng2025theoretical,wu2025fast}. A widely adopted remedy is to unmask only high-confidence tokens at each iteration~\citep{chang2022maskgit,gong2024scaling,nie2025large,wu2025fast}. These confidence-based heuristics partially mitigate the tension between decoding accuracy and throughput by approximating the true joint distribution. However from an information theoretical viewpoint, high probability means low information, thus prioritizing high-confidence tokens reduces the information revealed per step. Given the output $\bx$ contains fixed total information $\log p(\bx)$, decoding strategies that prioritize only the highest-confidence tokens should require more iterations than approaches that do not depend solely on this heuristic when unmasking tokens. This argument suggests a relationship between the computational cost (number of decoding rounds) and total information $\log p(\bx)$, and thus gives rise to the first question:
\begin{center}
    \emph{Can the computational cost of decoding be characterized in terms of fundamental information-theoretic quantities?}
\end{center}

If such a characterization exists, it would naturally suggest a design principle: \textit{to minimize decoding rounds, we should maximize the information decoded per round}. This can be achieved through two complementary directions. First, confidence-based methods inevitably unmask low-confidence tokens when there are no high-confidence tokens available. Empirically, we observe that these low-confidence tokens contribute substantial information to the overall decoding trajectory (see Figure~\ref{fig:exploration efficiency}). Moreover, some low-confidence tokens (e.g., variable names in coding) are intrinsically more informative than others because unmasking them unlocks cascades of high-confidence tokens in subsequent decoding iterations (e.g., all later references must use the same variable name). Second, existing block-based decoding strategies process blocks strictly sequentially. This causes the final few tokens in each block to be decoded essentially one by one rather than simultaneously with tokens in future blocks, thereby limiting the \emph{number of tokens} that can be decoded per round. Existing approaches lack principled mechanisms to address either limitation. This gap motivates our second question:
\begin{center}
    \emph{How can we systematically maximize the information decoded per round, both by identifying high-information tokens and by expanding parallel decoding opportunities?}
\end{center}

\begin{figure}
\centering
\begin{subfigure}{\linewidth}
    \centering
    \includegraphics[width=\linewidth]{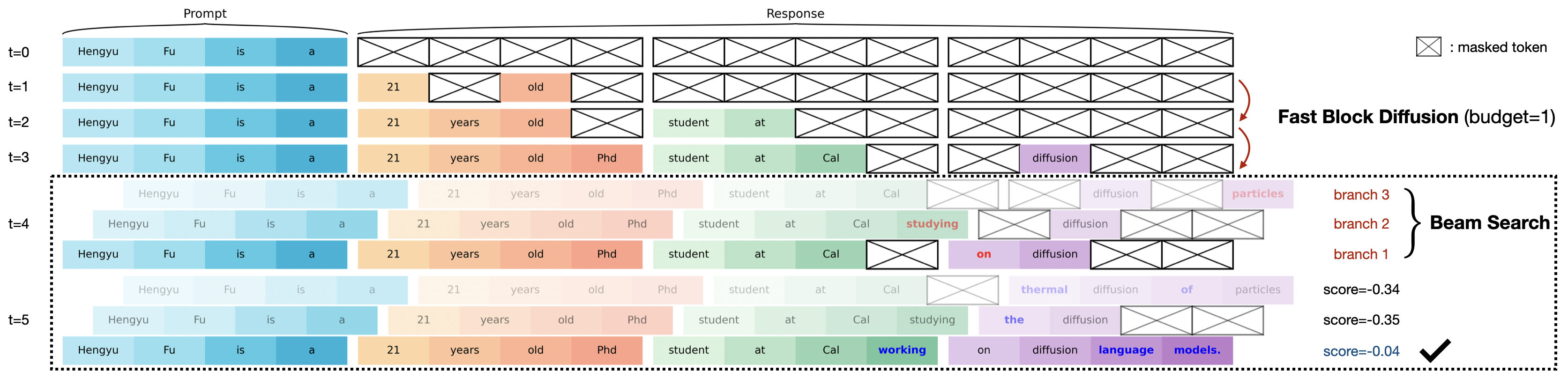}  
    \caption{Illustration of \fullname{} (\name{}): Fast block diffusion scheme proceeds by sequentially unlocking a new block to the right and applying confidence-based parallel sampling to all past blocks. At $t=4$, targeted exploration is triggered to apply beam search on 3 candidate exploration tokens (in red). At the next step, several high confidence tokens (in blue) are unlocked, and branch 1 with the highest score is committed}
  \end{subfigure}\hfill
  \begin{subfigure}{\linewidth}
    \centering
    \includegraphics[width=0.8\linewidth]{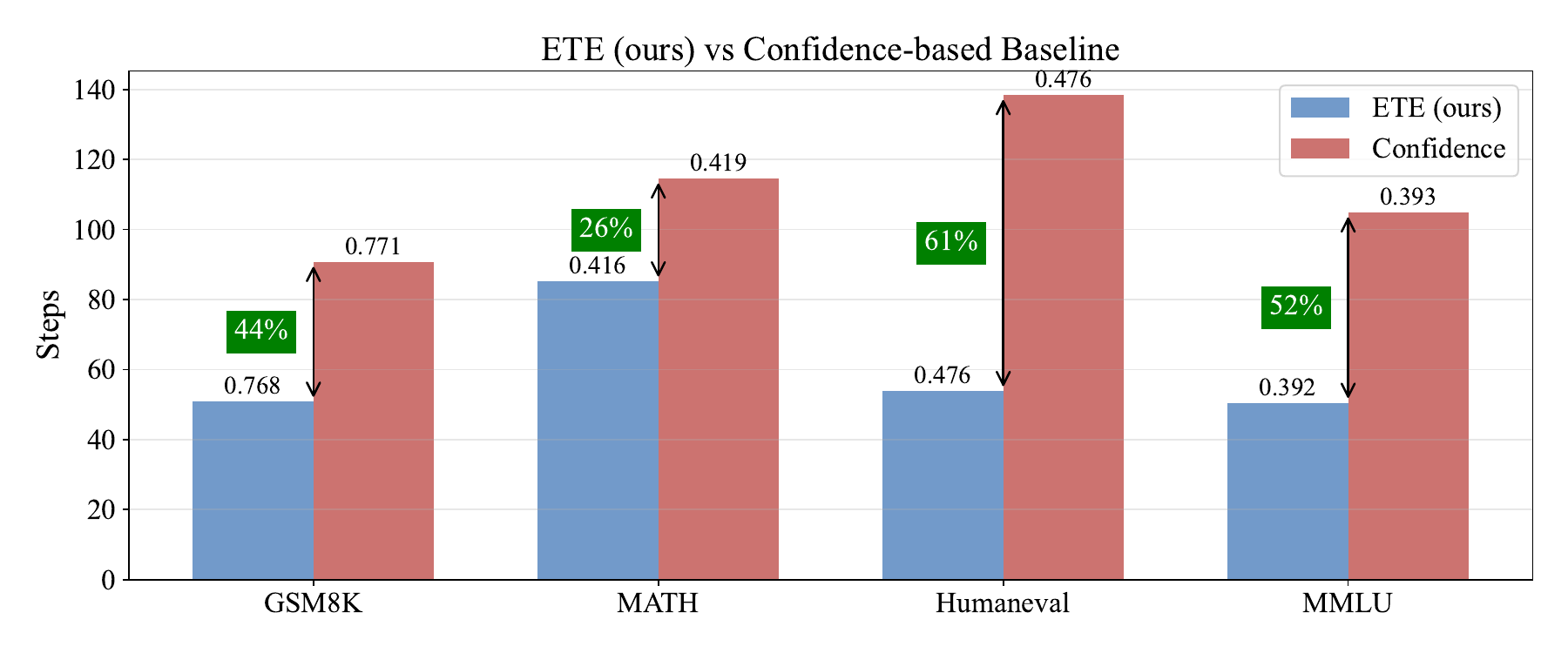}  
    \caption{Superior performance of \name{} across four standard benchmarks. With generation length fixed at 512 for MATH, GSM8K, and HumanEval, and 256 for MMLU-Pro, ETE (our algorithm) matches the accuracy of the confidence-based baseline~\citep{wu2025fast} while reducing inference steps by 26-61\% (accuracy values are annotated directly on each bar). The figure highlights the high-accuracy regime, where both methods attain similar accuracy under the vanilla block-diffusion LLaDA sampler \citep{nie2025large} which decodes one token per step. Section \ref{sec::experiment} details the experimental setup, and Figure \ref{fig:frontiers} shows the full accuracy-step trade-off.}
  \end{subfigure}
\label{fig:illustration}  
\end{figure}
\paragraph{Main Contributions}
In this paper, we formally address the above questions. Our main contributions are twofold:
\begin{itemize}
    \item \textbf{Information-theoretic lower bound.} We formalize the relationship between computational cost and information content by proving a lower bound on the number of decoding rounds:
$$
R \ge \max \left(\frac{-\log p(\bx)}{\log \left(\frac{n+1}{(1-f)n+1}\right)}, \frac{-\log p(\bx)-\epsilon}{f}\right),
$$
where $n$ is the sequence length, $\epsilon$ is the total approximation error of parallel decoding, and $f$ is the confidence threshold factor ensuring $(1+|A_r|)(1-p(x^i\mid x_{C_r})) \le f$ for all unmasked positions $i \in A_r$ in round $r$ given the previously unmasked positions $C_r$. This bound reveals a fundamental structure: 
$$
\text{Rounds} \ge \frac{\text{Total Information}}{\text{Information per Round}}.
$$
Reducing confidence threshold $f$ (stricter requirements) or approximation error $\epsilon$ shrinks the denominator, forcing more rounds. 

\item \textbf{Exploration-aware decoding algorithm with cross-block decoding.} Motivated by our theoretical insights and empirical observations, we introduce \fullname{} (\name{}), a new algorithmic framework that systematically maximizes information decoded per round through two complementary mechanisms:
\begin{itemize}
    \item \emph{Fast block diffusion sampling}: We extend standard block diffusion by enabling cross-block parallel decoding. Rather than processing blocks strictly sequentially, we assign a fixed step budget per block and permit high-confidence tokens in earlier blocks to be unmasked in parallel with the current block. This design expands the feasible set of tokens that can be decoded per round, directly increasing the numerator of "tokens per round" in our framework.
    
    \item \emph{Strategic exploration mechanisms}: We propose two  strategies for identifying and decoding high-information tokens: (i) \emph{inter-block implicit exploration}, which proactively decodes the highest-confidence token within "unsure" blocks in parallel across active blocks, ensuring forward progress; and (ii) \emph{targeted exploration via beam search}, which uses look-ahead search over medium-confidence positions to identify tokens that unlock the largest cascades of high-confidence decisions. 
\end{itemize}

\end{itemize}

Together, these mechanisms directly amplify the denominator in our lower bound by increasing both the information per token and the number of tokens decoded per round. Across four standard benchmarks (MATH, GSM8K, HumanEval, MMLU-Pro), \name{} consistently outperforms strong confidence-based baselines in both decoding efficiency and output quality. We further identify a \emph{free-lunch regime} where beam search introduces negligible wall-clock overhead, enabling exploration at minimal practical cost. These results underscore the importance of principled exploration in diffusion decoding and demonstrate a promising path toward closing the gap between parallel DLM inference and the joint distribution it aims to approximate.

\subsection{Related works}

\paragraph{Diffusion Language Models}
Discrete diffusion models have been widely adopted for categorical data generation, including natural language \citep{nie2025large}, biological sequences \citep{sahoo2024simple}, code synthesis \citep{singh2023codefusion}, and audio generation \citep{yang2023diffsound}. The framework was first proposed by \citet{NEURIPS2021_958c5305} as a discrete variant of Denoising Diffusion Probabilistic Models (DDPM) \citep{ho2020denoising}, and later reformulated as a probability ratio learning problem \citep{lou2023discrete}. Early models typically operated at small scales (fewer than 1B parameters). Through reparameterization \citep{zheng2023reparameterized} and engineering efforts such as KV-caching \citep{liu2025dllm}, masked (absorbing) diffusion models have emerged as the predominant paradigm due to their scalability. This scalability has enabled the development of diffusion-based large language models (DLMs) \citep{nie2025large,ye2025dream,zhu2025llada,song2025seed}, which achieve performance and inference speed comparable to autoregressive (AR) language models \citep{prabhudesai2025diffusion}. Recent works have also explored hybrid approaches through block-wise diffusion \citep{chen2024diffusion,arriola2025block}, combining the flexible content/length generation of AR models with the parallel inference capabilities of DLMs.

\paragraph{Parallel Decoding in DLMs}
Unlike autoregressive language models that generate tokens sequentially from left to right, DLMs frame generation as an iterative denoising (unmasking) process over entire sequences, and enable parallel decoding through bidirectional attention. Early approaches adopted fixed-step parallel decoding \citep{chang2022maskgit,nie2025large}, while more recent methods introduce adaptive strategies. \citet{yu2025dimple} propose confident decoding, which dynamically unmasks high-confidence tokens above a fixed threshold. \citet{wu2025fast} extends this with a dynamic confidence threshold mechanism, and \citet{wei2025accelerating} introduces a two-phase approach, which alternates between two decoding stages, performing conservative decoding until high-confidence regions form and then conducting aggressive confidence-based parallel decoding on those confident regions. \citet{huang2025pc} addresses positional bias through confidence calibration, while \citet{ben2025accelerated} leveraged entropy-based unmasking with controlled approximation error. \citet{kim2025train} introduced using probability margin as the confidence measure. Additional acceleration techniques include attention optimization via caching mechanisms \citep{liu2025dllm} and sparsification \citep{song2025seed}. However, all these approaches share a common principle: prioritizing certain (high-confidence/low-entropy) tokens first. This strategy inherently reduces the information revealed per inference step, potentially limiting overall decoding speed from an information-theoretic perspective, given that the total information content is fixed. Notably, recent findings in Reinforcement Learning with Verifiable Rewards (RLVR) suggest that uncertain (high-entropy) tokens are key to successful and efficient reasoning \citep{wang2025beyond}, which contrasts with the current certain-token-first paradigm in DLM decoding. A concurrent work~\citet{lee2025lookahead} employs lookahead generation guided by verifier rewards to identify the correct decoding trajectory. In contrast, our usage of lookahead generation restricts to targeted exploration, with the primary goal of unlocking more confident tokens and thereby accelerating inference, instead of higher accuracy.

\section{Background}

\paragraph{Notations} We use bold symbols $\bx = (x^1,\dots,x^N)$ to represent a sequence of $N$ tokens. We also use the standard big-O notations: $O(\cdot)$ and $\Omega(\cdot)$ to hide absolute positive constants. Let $p^i$ stand for the marginal distribution at position $i$ induced by the sequence distribution $p$: when the position is clear from context, we write $p$ for brevity. 

\paragraph{Masked Diffusion Models.}

Masked Diffusion Models (MDMs) are a subclass of discrete diffusion models. By extending the vocabulary $\V$ with a mask token $\mask$, MDMs progressively transform a clean sequence $\bx_0 = (x_0^1, \dots, x_0^N)$ into a fully masked sequence $(\mask,\dots,\mask)$ in the forward process following an absorbing noising schedule:
\begin{align*}
    q_{s|0}(\bx_s \mid \bx_0) = \prod_{i=1}^N q_{s|0}(x_s^i \mid x_0^i), \quad q_{s|0}(x_s^i \mid x_0^i) = \begin{cases}
        \alpha_t, &~ x_s^i = x_0^i,\\
        1-\alpha_t, &~ x_s^i = \mask.
    \end{cases}
\end{align*}
At training time, a parametric model $p_\theta$ is trained to take $\bx_s$ as input and predict all masked tokens simultaneously, by fitting the ELBO loss~\citep{shi2024md4,ou2024,sahoo2024,nie2025large}:
\begin{align*}
    \mathcal{L}(\theta) = -\mathbb{E}_{s,x_0,x_s}\left[\frac{1}{s}\sum_{i=1}^N \mathbbm{1}(x_s^i = \mask) \log p_\theta(x_0^i\mid \bx_s) \right].
\end{align*}
Thus, given a partially unmasked tokens $\bx_s$, $p_{\theta}(\cdot|\bx_s)$ models the \textbf{conditional marginal distribution} of all masked positions given the unmasked part of $\bx_s$. Moreover, this ELBO loss can be reformulated into a time-agnostic version~\citep{zheng2024masked}: 
\begin{align*}
    \mathcal{L}(\theta)=-\mathbb{E}_{\sigma \sim {\rm Unif}(S_N)}\left[\sum_{i=1}^{N}\sum_{k \in \sigma(\ge i)}\frac{1}{N-i+1}\log p_{\theta}(x^{k}_0|\bx_{\sigma(<i)})\right],~~x^{j}_{\sigma(<i)}=\left\{\begin{aligned}
        &x_0^{j},~~~j\in \sigma(<i)\\
        &\mask,~~~ j\in \sigma(\ge i)
    \end{aligned} \right.
\end{align*}
where $\sigma$ is a random permutation sampled from the uniform distribution over the set $S_N$ of all permutations on $[N]$, and $\bx_{\sigma(<i)}$ represents a partially masked version of the clean sequence $\bx_0$, with indices in $\sigma(\ge i)$ masked. This demonstrates that $p_{\theta}$ intrinsically parametrizes an \textbf{any-order autoregressive model}, able to predict the entire sequence with any specified generation order $\sigma$.

\paragraph{Inference of Masked Diffusion Models.}

At inference time, MDMs discretize the reverse process by iteratively reconstructing clean sequences from a fully masked state. Via a Gillespie-style sampler~\citep{gillespie1976general,gillespie1977exact,peng2025path}, the reverse process can be simplified into a time-independent, uniform-order denoising process equivalent to an any-order autoregressive generation~\citep{ou2024}. 

Specifically, we start with a fully masked sequence $\bx_0=[\mask^{\otimes N}]$ and randomly sample a generation order $\sigma \sim {\rm Unif}(S_N)$. At step $t\in[N]$, we unmask position $\sigma(t)$ by sampling from $x^{\sigma(t)}\sim p(x^{\sigma(t)}|\bx_{t-1})$ and let $\bx_{t}$ be $\bx_{t-1}$ with position $\sigma(t)$ replaced by $x^{\sigma(t)}$. In this process, a token remains fixed for the remainder of the denoising steps once it is unmasked, and this token-by-token random-order sampler is theoretically consistent with the ELBO reformulation..

However, this vanilla approach faces two key limitations. First, masked diffusion models cannot handle flexible sequence lengths since the encoder-type architecture requires a predefined context length. Second, it offers no efficiency advantages over traditional autoregressive counterparts, because it decodes only one token per step while requiring more computation due to its bidirectional self-attention architecture and lack of KV-cache support.

To address these limitations, two key strategies have been proposed. To enhance flexibility in generation length, \textbf{block diffusion sampling}~\citep{arriola2025block,nie2025large} partitions the response window into consecutive blocks and performs block-wise semi-autoregressive generation. To improve efficiency and fully exploit the bidirectional attention mechanism, \textbf{parallel decoding}~\citep{nie2025large,wu2025fast} strategically selects a generation order instead of a random one and compresses the generation process by decoding multiple tokens in a single round. We briefly introduce these two strategies below.

\begin{figure}[h]
  \centering
  \begin{subfigure}{0.50\textwidth}
    \centering
    \includegraphics[width=\linewidth]{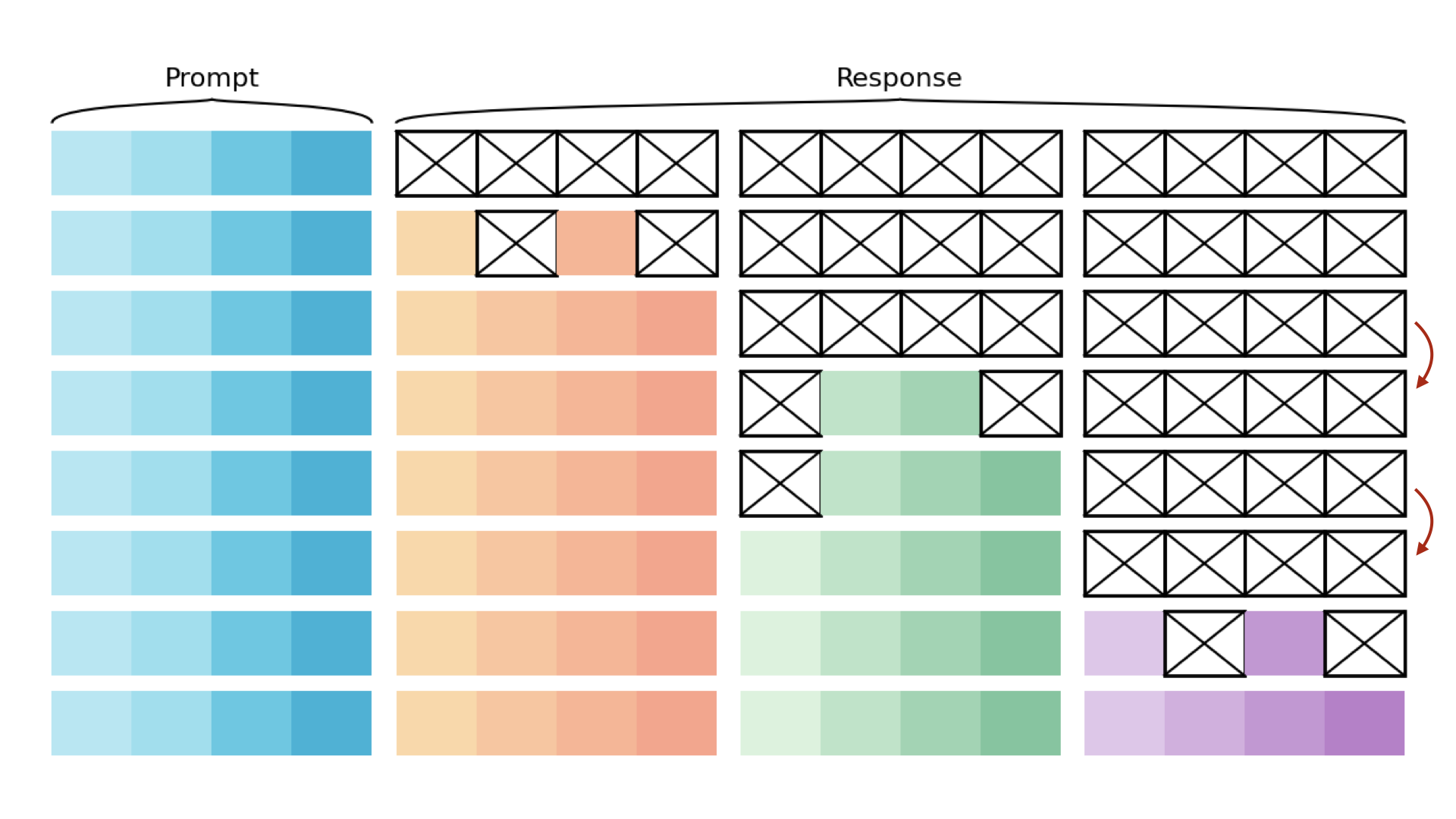}
    \caption{Block diffusion LLaDA sampling~\citep{nie2025large}.}
  \end{subfigure}\hfill
  \begin{subfigure}{0.50\textwidth}
    \centering
    \includegraphics[width=\linewidth]{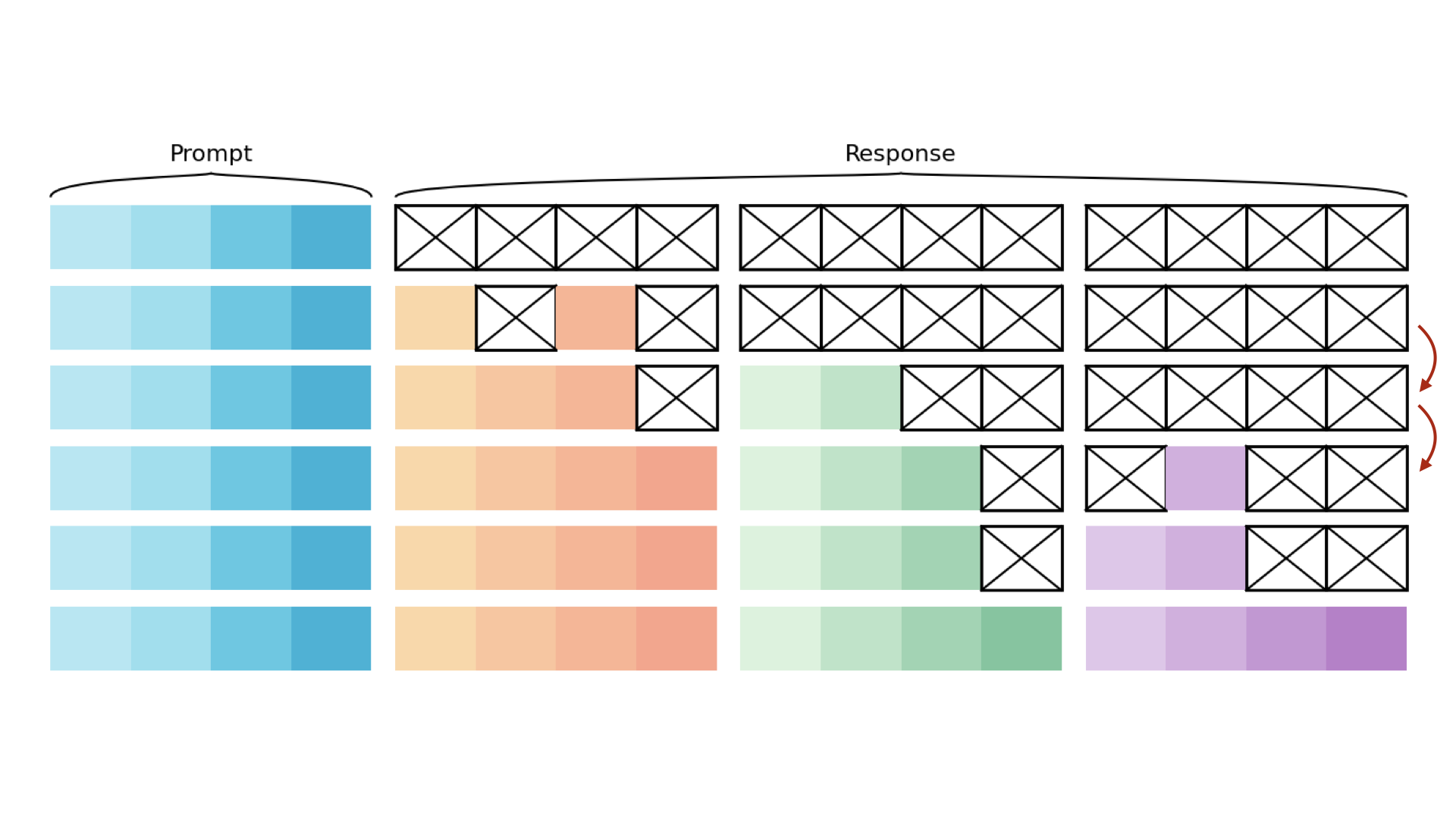}
    \caption{Fast block diffusion sampling.}
  \end{subfigure}
  \caption{Panel (a): Block diffusion unlocks the next block after the current block is fully unmasked; Panel (b): Fast block diffusion (with budget 1) unlocks the next block after the budget exhausts and retains the ability to unmask prior blocks, enabling faster decoding. In both figures, we use red arrows to indicate unlocking the next block.}
  \label{fig:fast-block-diffusion}
\end{figure}

\paragraph{Block Diffusion Sampling} 
Given a prompt sequence $\bx_{\rm prompt}$ of length $n_0$ and a target generation length $n$, the sequence is initialized as 
$\bx_0=[\bx_{\rm prompt},\mask^{\otimes n}]\in \mathcal{V}^{n+n_0}$, which concatenates the prompt (as a prefix) with a response window of $n$ mask tokens. Given a block length $n_b$, the response window is partitioned into $L = n/n_b$ consecutive blocks $\mathcal{B}_b = \left[n_0 + {(b-1) n_b}: n_0 + {b n_b}\right]$ $(b=1,\dots,L)$, each of length $n_b$. Tokens are then decoded block-by-block from left to right. Panel (a) of Figure~\ref{fig:fast-block-diffusion} provides a visualization of the block diffusion sampling process.

\paragraph{Confidence-based Parallel Decoding} 
Parallel decoding aims to unmask multiple tokens simultaneously in a single inference step. At each decoding step $t$ within block $b$, given the partially unmasked sequence $\bx_t$, the model first generates predictions at all masked positions in the block via greedy parallel decoding of their conditional marginals:
\begin{align*}
  \hat x_{t+1}^i  = \arg\max_{v \in \V} p_\theta^i(v \mid \bx_t),~~~i\in  S:=\{i:x_t^i=\mask ~\text{and index $i$ is in block $b$}\}
\end{align*}
Since the product of conditional marginal distributions over all masked tokens may differ substantially from the true conditional joint distribution, parallel-decoding strategies selectively commit only a subset of predictions to ensure small approximation error. As the predominant strategy in parallel decoding, confidence-based parallel-decoding methods selectively unmask tokens in block $b$ based on their confidence scores: $c^i = p^{i}_\theta(\hat x_{t+1}^i \mid \bx_t)$. We describe three variants of confidence-based decoding strategies:
\begin{itemize}
   \item[(1)] \textbf{Fixed-number scheme:} Given a fixed number $k>0$ we expect to decode per round, \citet{nie2025large} unmask the top-$k$ confident tokens within the block.
   \item[(2)] \textbf{Static confidence threshold:} \citet{yu2025dimple} propose a static threshold scheme that unmasks all tokens exceeding a fixed confidence threshold $C \in (0,1)$.
   \item[(3)] \textbf{Dynamic confidence threshold:} \citet{wu2025fast} extend the static threshold strategy to a dynamic variant that sorts confidences as $c^{(1)} > \dots > c^{(m)}$ and determines the maximum integer $k$ satisfying $(k + 1)(1 - c^{(k)}) < f$, where $f > 0$ is a predefined threshold. The algorithm then unmasks the corresponding top-$k$ confident tokens $\hat x_{t+1}^{(1)},\dots, \hat x_{t+1}^{(k)}$.
\end{itemize}
Regardless of which specific selection strategy is used, we refer to this step as an \textbf{``exploitation''} step, as it prioritizes and exploits high-confidence predictions. Intuitively, tokens are nearly \textbf{conditionally independent} when they are highly confident, which means the product of their marginals closely approximates their joint distribution. Notably, for the latter two confidence-based decoding strategies, decoding would stall when the current block contains no tokens exceeding the confidence threshold. To prevent this, the highest-confidence token is unmasked regardless of whether its confidence exceeds the threshold. We refer to this as an \textbf{``implicit exploration''} step to distinguish from the exploitation step, since an implicitly low-confidence token is  decoded in this step. We provide pseudocode for the entire confidence-based decoding process with a static confidence threshold $C$ in Algorithm~\ref{alg:baseline}.

\begin{algorithm}[H]
\caption{Confidence-based parallel decoding with block diffusion sampling}
\label{alg:baseline}
\begin{algorithmic}[1]
\State \textbf{Input:} Model $p_\theta$, prompt $\bx_{\text{prompt}}$, generation length $n$, num blocks $L$, confidence threshold $C$.
\State $\bx_0 \leftarrow \left(\bx_{\text{prompt}}, \mask^{\otimes n}\right), ~t \leftarrow 0$ \Comment{Initialize sequence and step counter}
\State $n_0 = |\bx_{\text{prompt}}|, ~\mathcal{B}_b = \left[n_0 + \frac{(b-1) n}{ L}: n_0 + \frac{b n}{ L}\right] ~(b=1,\dots,L)$
\For{$b = 1 \to L$}\Comment{Iterate over blocks}
    \While{block $b$ is not fully unmasked} 
    \State $S \leftarrow \{0 \leq i \leq n_0+n: x_t^i = \mask\} \cap \mathcal{B}_b$ \Comment{Only unmask the current block}
    \State $\hat x_{t+1}^i \leftarrow \arg\max_{v \in \V} p_\theta^i(v \mid \bx_t), ~i \in S$ \Comment{Greedy decoding}
    \State $x_{t+1}^i \leftarrow \begin{cases}
\hat x_{t+1}^i, & \text{if } p^{i}_\theta(\hat x_{t+1}^i \mid \bx_t) > C ,\\
x_t^i, & \text{otherwise}.
\end{cases}$
        \Comment{Exploitation}
        \If{$\bx_{t+1} = \bx_t$} 
            \State $i^*, \hat x_{t+1}^{i^*} \leftarrow \arg\max_{i \in S, v \in \V} p_\theta^i(v \mid \bx_t), ~ x_{t+1}^{i^*} \leftarrow \hat x_{t+1}^{i^*}$
            \Comment{{Implicit Exploration}}
        \EndIf
        \State $t \leftarrow t+1$
    \EndWhile
\EndFor
\State \textbf{Return} $\bx_t$
\end{algorithmic}
\end{algorithm}

\section{Inefficiency of Confidence-based Decoding: counterexamples and empirical observations}\label{sec::inefficiency}

While exploitation steps form the core of current confidence-based decoding, they face fundamental limitations that need investigation. Before formalizing our theoretical framework, we present two illustrative examples that highlight a critical insight: high-confidence tokens can be locally easy to predict yet globally uninformative, while low-confidence tokens, if decoded early, can unlock substantial parallelism. 

\paragraph{Example 1: Structured Profile Records}
Consider randomly sampling the profile of an undergraduate student from a structured database:
\begin{align*}
    \bx =[\texttt{name},\texttt{age},\texttt{school},\texttt{hobby}]\sim p_{\text{data}}.
\end{align*}
A confidence-based decoding algorithm typically selects the field \texttt{age} first, since its predicted distribution is concentrated within a narrow range (18--22), yielding artificially high confidence. 
However, if one decodes \texttt{name} first---despite its relatively diffuse predictive distribution---the remaining fields often become nearly deterministic: the student's \texttt{school}, \texttt{age}, and \texttt{hobby} are strongly constrained once the identity is known.  
Thus, a naive ``confident-token-first'' decoding strategy is locally optimal but globally inefficient: it picks the token that resolves the \emph{least} uncertainty.

\paragraph{Example 2: Code Generation}
A similar phenomenon arises in coding tasks. Consider predicting the following masked snippet for computing a factorial:
\begin{verbatim}
def factorial(n):
    [MASK] = 1
    for i in range(1, n + 1):
        [MASK] *= i
    return [MASK]
\end{verbatim}
Each possible masked variable name (e.g., \texttt{ans}, \texttt{result}, \texttt{res}) may have low marginal conditional probability due to the arbitrary naming convention, resulting in low-confidence among all the three masked tokens. 
Nevertheless, decoding \emph{any one} of the masked slots immediately forces the remaining masks to adopt the same name, thereby collapsing the search space for all subsequent predictions.  
Again, high-entropy tokens, if decoded strategically, can unlock large determinism elsewhere.

\begin{figure}[t]
\centering
\includegraphics[width=\linewidth]{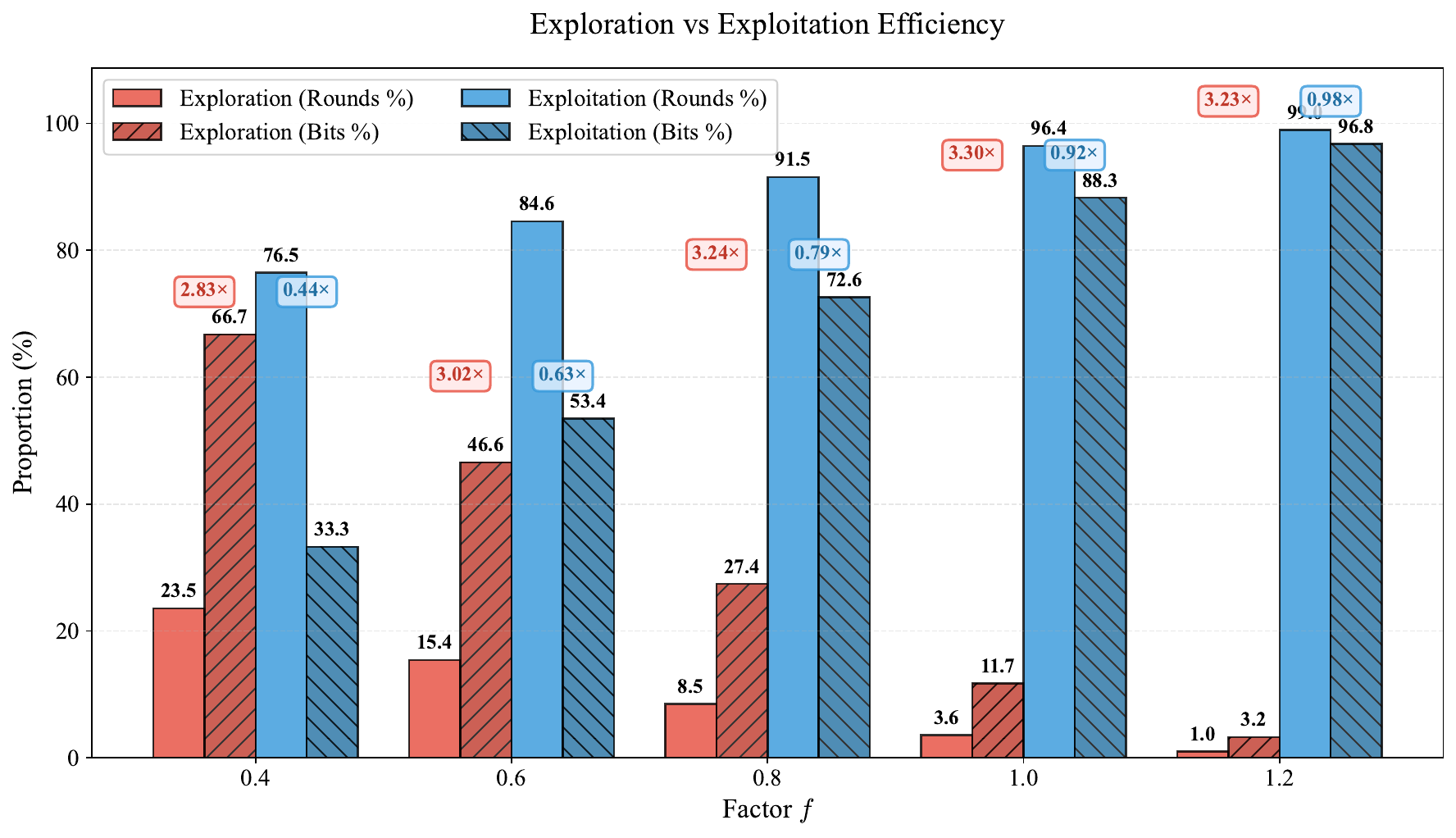}  
\caption{Side-by-side comparison of (implicit) exploration and exploitation across rounds distribution and bits distribution for each factor $f$. For each factor, four bars show: exploration rounds \% , exploration bits \% , exploitation rounds \%, and exploitation bits \%, with each proportion computed by averaging over the same 200 samples from GSM8K dataset. Colored boxes display efficiency ratios (Bits \% / Rounds \%) for exploration (red) and exploitation (blue). }
\label{fig:exploration efficiency}
\end{figure}

\paragraph{Information contribution of exploration and exploitation}

To quantify the inefficiency of parallel decoding, we evaluate the dynamical confidence-aware decoding strategy~\citep{wu2025fast} across different confidence factors on the GSM8K dataset~\citep{cobbe2021gsm8k}, with the detailed experimental setup  deferred to Section \ref{sec::experimentsetup}. Figure~\ref{fig:exploration efficiency} shows that exploration achieves an efficiency ratio of approximately 2.8--3.3x (bits proportion per rounds proportion), while exploitation yields only 0.44--0.98x across different confidence factors. At the most conservative setting ($f=0.4$), exploration contributes 66.7\% of total bits using only 23.5\% of rounds, whereas exploitation consumes 76.5\% of rounds to contribute merely 33.3\% of bits. This dramatic efficiency gap emerges because (implicit) exploration targets high-uncertainty positions, whose decoding cascades into large information gains. In contrast, exploitation refines already-confident predictions, yielding negligible additional information. This empirical observation demonstrates that confidence-based decoding is fundamentally inefficient in revealing new information because it prioritizes certainty over information throughput.

\paragraph{Key Observation} From the intuitive examples and empirical analysis above, a central phenomenon emerges: confidence-based decoding heuristics prioritize high-confidence positions despite their minimal global information contribution. However, in structured modalities such as code, profiles, tables, and graphs, low-confidence tokens frequently carry disproportionate information because resolving their uncertainty strongly constrains downstream positions. This leads to our core insight:

\begin{center}
\emph{Decoding \textbf{highly informative tokens} can vastly accelerate generation, as they often resolve many dependent positions simultaneously.}
\end{center}

This observation together with the two illustrative examples motivate two directions: (i) theoretically analyzing the fundamental limits of confidence-constrained decoding algorithms, and (ii) empirically designing principled exploration strategies that identify and target high-information yet low-confidence tokens.

\subsection{Information-Theoretic Lower Bounds of decoding steps}

\paragraph{Setup and Notation}
Let $p(\cdot)$ be a distributional model over length-$n$ sequence that models the conditional probability $p(x^A|x^B)$ for any disjoint subset $A, B$ of $[n]\equiv \{1,\ldots,n\}$. 
A \emph{parallel decoding schedule} is an ordered partition of indices
\[
A_1 \cup\, A_2 \cup\, \cdots \cup\, A_R = [n], \quad A_i \cap A_j = \emptyset
\]
such that $x^{A_r}$ is the set of tokens unmasked in rounds $r$ for $r=1,\ldots,R$. At the start of round $r$, the set of unmasked indices is $C_r \triangleq \bigcup_{s<r} A_s$, and the mask indices are $U_r \triangleq [n]\setminus C_r$. A parallel decoding algorithm will select the subset $A_r \subseteq U_r$ to be decoded in this round based on the conditional marginals $p(x^i|x^{C_r})$ over all mask indices $i\in U_r$. Due to parallel decoding, an approximation error will arise due to the gap between the true conditional probability $p(x^{A_r}\mid x^{C_r})$ and the actual sampling probability $\prod_{i\in A_r} p(x^i\mid x^{C_r})$. We let
\begin{align*}
    \epsilon_r=   \left|-\log p(x^{A_r}\mid x^{C_r}) - \sum_{i\in A_r} \bigl[-\log p(x^i\mid x^{C_r})\bigr]\right|
\end{align*}
be the approximation error in log probabilities in round $r$. A small $\epsilon_r$ ensures that the conditional joint likelihood of $x^{A_r}$ given $x^{C_r}$ is close to the product of the conditional marginals of $x^{C_r}$, which allows parallel decoding with small approximation error. Also, we define the total approximation error by
\begin{align}\label{eq:epsilon}
    \epsilon=\left|-\log p(\bx) - \sum_{r=1}^{R}\sum_{i\in A_r} \bigl[-\log p(x^i\mid x^{C_r})\bigr]\right|,
\end{align}
which is upper bounded as $\epsilon\le \sum_{r=1}^{R}\epsilon_r$ by the chain rule of probability $p(\bx)=\prod_{r=1}^{R}p(x^{A_r}\mid x^{C_r})$ and the triangle inequality.

In confidence-based methods, token $x^i$ is unmasked if and only if its confidence $p(x^i\mid x^{C_r})$ satisfies a lower bound. Therefore, we work on the following assumption captures a broad family of practical confidence-based rules used in DLLM decoding (e.g., \cite{wu2025fast}). 
\begin{assumption}[Dynamic Threshold Confidence Decoding]
\label{asp:dynamic_threshold}
The selection rule for each round $r$ ensures that every chosen position $i\in A_r$ satisfies
\begin{align*}
(1+|A_r|)(1-p(x^i\mid x^{C_r})) \;\le f,~~~f\le 1.
\end{align*}
\end{assumption}

The quantity $1-p(x^i\mid x^{C_r})$ measures the algorithm's uncertainty about position $i$, and the factor $(1+|A_r|)$ enforces that as more tokens are decoded jointly in round $r$, each token must individually meet a stricter confidence threshold. This assumption also theoretically guarantees that greedy parallel decoding yields the same result of greedy sequential decoding on $A_r$ (Theorem 1, \citet{wu2025fast}). 
Under this assumption, we prove an explicit lower bound on the number of rounds required by \emph{any} decoding algorithm obeying this dynamic confidence constraint.

\begin{theorem}[Step Lower Bound for Confidence-Based Parallel Decoding]
\label{thm:main}
Consider any parallel decoding schedule and any length-$n$ sequence $\bx=(x^1,\ldots,x^n)$ satisfying Assumption~\ref{asp:dynamic_threshold}, and let the total approximation error $\epsilon$ be defined in Eq.~\eqref{eq:epsilon}. Then the number of rounds $R$ must satisfy
\begin{align}
\label{eq:lowerbound}
 R \ge \max \left(\frac{-\log p(\bx)}{  \log \left(\frac{n+1}{(1-f)n+1}\right)},\frac{-\log p(\bx)-\epsilon}{ f}\right)
\end{align}
\end{theorem}

\paragraph{Interpretation.}
Theorem~\ref{thm:main} formalizes a fundamental trade-off in confidence-based parallel decoding. The lower bound explicitly couples the required number of rounds $R$ to two quantities: it is proportional to the total amount of information in the sequence ($-\log p(\bx)$) and inversely proportional to the per-round information limit $\log \left(\frac{n+1}{(1-f)n+1}\right)$ determined by the confidence factor $f$. Moreover, if taking the approximation error $\epsilon$ into account, we need to decode at least $-\log p(\bx)-\epsilon$ \textit{valid bits} in total. We can show a stricter argument that parallel decoding algorithm can only provide $f$ valid bits in each round, requiring at least ${(-\log p(\bx)-\epsilon)}/{ f}$ rounds in total to achieve an $\epsilon$ approximation error. Thus, as $f$ decreases, which means the decoder imposes stricter confidence requirements, the denominator shrinks, forcing $R$ to grow faster with sequence information. Critically, this dependence holds regardless of the underlying distribution's structure, even when the data exhibits strong conditional independencies that could enable faster parallel decoding. The proof is deferred to Appendix \ref{sec:proof}.

Particularly, when the total amount of information $-\log p(\bx)= \Omega{(n)}$, since per-round information limit is at most $O(\log n)$ bits, we need essentially at least $\Omega{(n/\log n)}$ rounds in total even regardless of the approximation error. This result reveals that confidence-based heuristics become inherently sequential in high-entropy regimes, unable to exploit the parallelism that structured dependencies provide. This structural inefficiency results from a fundamental mismatch: confidence-based selection diminishes information gain, instead greedily resolving certainty. Thus, overcoming this limitation requires decoding strategies that \emph{actively identify and resolve high-entropy positions}, transforming them into conditional constraints that unlock parallelism elsewhere. 

\section{\name{}: \fullname{}}

\subsection{From Theory to Algorithm: The Golden Principle}

Our information-theoretic analysis in Section~3 reveals a simple but fundamental relationship:
\begin{align}
\text{Rounds} \geq \frac{\text{Total Information (Bits)}}{\text{Information per Round (Bits per Round)}}.
\end{align}
This inequality exposes the core inefficiency of confidence-based decoding: by prioritizing high-confidence (low-information) tokens, these methods may decrease the denominator (the information decoded per round), thereby forcing a larger number of rounds overall.

This insight establishes our algorithmic design principle: \emph{to minimize decoding rounds, we must maximize the information decoded per round}. We achieve this through a two-fold approach:

\begin{itemize}
    \item \textbf{Expanding the decoding canvas (Fast Block Diffusion):} By enabling parallel decoding across multiple blocks simultaneously, we increase the \emph{number of tokens} that can potentially be decoded in each round, raising the opportunities for high-information contribution each round.
    
    \item \textbf{Targeting high-information tokens (Strategic Exploration):} By explicitly identifying and decoding high-entropy tokens--those that carry the most information and unlock cascades of downstream predictions--we increase the \emph{information per token} decoded in each round.
\end{itemize}

Together, these mechanisms directly increase the denominator in our lower bound, breaking through the inherent limitations of confidence-based methods. In the remainder of this section, we detail each component and show how they combine to form \name{} (\fullname{}), a principled algorithm that operationalizes these information-theoretic principles.

\subsection{Fast Block Diffusion Sampling: Expanding the Decoding Canvas}\label{sec:fast-block-diffusion}

\paragraph{Motivation: expanding parallel decoding range.} Standard block diffusion processes blocks sequentially, unlocking block $b+1$ only after block $b$ is completely decoded. This rigid left-to-right constraint artificially limits parallelism and under-utilize the bi-directional attention mechanism: high-confidence tokens in earlier blocks cannot be decoded while processing later blocks, even when doing so would provide valuable bidirectional context. To increase the number of tokens decoded per round, we need to break this sequential bottleneck.

 Following block diffusion LLaDA sampling~\citep{nie2025large,wu2025fast}, \name{} partitions the sequence $\bx$ of length $n$ into $L$ contiguous blocks $\mathcal{B}_1, \dots, \mathcal{B}_L$, with each block's length being $n_b=n/L$. Crucially, \name{} introduces two key modifications:

\begin{enumerate}
    \item \emph{Budget-based progression}: Rather than waiting for complete block decoding, we assign a uniform sampling budget of $N$ diffusion iterations per block and unlock the next block once this budget is exhausted. This prevents the algorithm from stalling in low-throughput regimes where blocks are nearly complete and few tokens remain to be decoded.
    
    \item \emph{Inter-block parallel decoding}: Unlike prior methods that restrict decoding to the current block, \name{} allows high-confidence tokens in \emph{earlier blocks} to be unmasked in parallel with the current block. This bidirectional information flow enables future positions to inform and confirm earlier predictions.
\end{enumerate}

Formally, let $b_t$ be the block index at step $t$ and $\M_t$ be the set of all masked positions after step $t-1$. The feasible positions for unmasking at step $t$ are:
\begin{align}\label{eq:feasible-pos}
    S_t = \M_t \cap \left(\cup_{b=1}^{b_t}\mathcal{B}_b\right),
\end{align}
which includes all masked positions in the current and all previous blocks. After exhausting the budget for the last block, we apply a few additional decoding rounds ($n_f$) to ensure later blocks are not under-decoded.

 By enabling cross-block parallel decoding, fast block diffusion increases the \emph{feasible set} of tokens that can be decoded in each round. High-confidence tokens that emerge in earlier blocks during later-block processing can now be immediately committed, extracting more information per inference step. This directly expands the numerator of "bits per round" in our theoretical framework.

\subsection{Confidence-based Exploitation}

\name{} retains the benefits of confidence-based token selection for exploitation. Specifically, let $\bx_t$ be the full sequence at step $t$, containing prompt tokens, unmasked tokens, and masked tokens. We run the diffusion LM once to obtain marginal probabilities $p_\theta^i(\cdot \mid \bx_t)$ for all positions $i \in S_t$. Following~\citet{yu2025dimple}, \name{} performs greedy decoding $\hat{x}_{t+1}^i=\arg\max_{v \in \V}p_\theta^i(v\mid \bx_t)$ and commits exploitation tokens whose confidence surpasses threshold $C>0$:
\begin{align}\label{eq:exploit-conf}
x_{t+1}^i = \begin{cases}
\hat x_{t+1}^i, & \text{if } c(\bx_t) > C \text{ and } i \in S_t,\\
x_t^i, & \text{otherwise},
\end{cases}
\end{align}
where $c^i(\bx_t) = p_\theta^i(\hat x_{t+1}^i| \bx_t)$ is the confidence at index $i$ given $\bx_t$.

\subsection{Exploration Mechanisms: Targeting High-information Tokens}

\paragraph{Motivation: increasing information per token.} While fast block diffusion expands \emph{how many} tokens can be decoded per round, it does not address \emph{which} tokens carry the most information. Our illustrative examples and empirical observations (Figure~\ref{fig:exploration efficiency}) in Section \ref{sec::inefficiency} demonstrate that some low-confidence, high-entropy tokens are precisely the positions that, when decoded, trigger cascades of newly confident predictions. However, confidence-based methods systematically avoid these high-information positions (unless when there are no confident tokens and implicit exploration is triggered). To maximize bits per round, we must strategically target and decode these tokens through principled exploration.

\name{} employs two complementary exploration mechanisms operating at different scales:

\paragraph{Inter-block implicit exploration.}
When a previous block lacks high-confidence tokens (none exceed threshold $C$), \name{} unmasks the highest-confidence masked token in that block. This ensures forward progress in every active block, preventing decoding from stalling. By interleaving exploration and exploitation tokens across blocks, \name{} can accelerate inference compared to block diffusion LLaDA. Empirically, this implicit exploration across blocks does not significantly degrade output accuracy, since tokens in different blocks exhibit lower correlation.

\paragraph{Targeted exploration via beam search.}
Beyond implicit exploration, \name{} performs principled exploration within the current block by strategically identifying and testing high-information tokens through look-ahead beam search.

\begin{itemize}
    \item \textbf{When to explore.} \name{} triggers targeted exploration when:
\begin{enumerate}
    \item The average confidence in the decoding frontier falls below threshold $\gamma$, which indicates an information-poor region where exploitation alone would require many rounds.
    \item The block has sufficient remaining masked tokens ($> N_e$) to justify the exploration overhead.
\end{enumerate}
This information-aware criterion ensures exploration focuses on genuinely difficult situations where the potential information gain justifies the computational cost.

\item \textbf{What to explore.} Rather than exploring uniformly, we target \emph{medium-confidence} positions near an information level $c^{\text{info}} \approx 0.2$. These positions represent genuine ambiguity (unlike near-certain tokens) yet have sufficient signal to resolve accurately (unlike near-zero confidence). Given beam size $k$, \name{} identifies exploration candidates via:
\begin{align}\label{eq:explore-space}
\mathcal{H}
= \text{Topk}_{i \in \text{current block} \cap \M_t}
\Bigl( - \bigl| c^i(\bx_t) - c^{\text{info}} \bigr| +\beta\cdot \left(i-(b_t - 1)\cdot n_b\right) \Bigr),
\end{align}
where the position bias $\beta>0$ slightly favors later tokens to exploit bidirectional context.

\item \textbf{Which to commit.} For each candidate $j \in \mathcal{H}$, \name{} creates a hypothesis $\bx_{t+1;j}$ by fixing position $j$ to $\hat x_{t+1}^j$ and performs \textbf{batched inference} on all $|\mathcal{H}|=k$ hypotheses simultaneously. This reveals how each exploration choice influences remaining masked positions. Each hypothesis is evaluated via:
\begin{align}
\label{eq:explore-score}
s(j) := \underbrace{\alpha\cdot \log c^j(\bx_t)}_{\text{sample quality}} ~+~  \underbrace{\log \sum_{i\in S_{t+1}\setminus\{j\}} c^i(\bx_{t+1;j}) \mathbf{1}\left(c^i(\bx_{t+1;j}) \geq C\right)}_{\text{induced high-confidence tokens}}.
\end{align}
The first term ensures the exploration token has good sample quality, while the second term measures how many downstream tokens become high-confidence after committing this choice. The hyperparameter $\alpha > 0$ balances these two objectives. \name{} identifies the candidate $j^{\star}=\arg\max_{j\in\mathcal{H}} s(j)$ achieving the highest score and commits the corresponding hypothesis sequence $\bx_{t+1;j^{\star}}$. After this exploration, one exploitation step can be performed directly to decode these induced high-confidence tokens.
\end{itemize}

Committing high-entropy tokens collapses diffuse distributions, triggering confidence cascades through conditional dependencies, which can be exploited immediately. Critically, the batched inference introduces no additional inference rounds--we perform two steps (exploration + exploitation) in what confidence-based methods would require for a single step. This may simultaneously increase both the information per token and tokens per round, directly amplifying the denominator in our lower bound.

\subsection{Algorithm Framework}
Integrating these components, Algorithm~\ref{alg:main} formalizes the complete \name{} sampling procedure, orchestrating exploitation and exploration phases within the fast block diffusion framework. Algorithm~\ref{alg:subroutines} details the constituent subroutines for exploitation, implicit exploration and targeted exploration, respectively.

\begin{algorithm}[h]
\caption{\textsc{\name{}} Main Sampling Procedure (Simplified)}
\label{alg:main}
\begin{algorithmic}[1]
\State \textbf{Input:} Model $p$, prompt $\bx_{\text{prompt}}$, sequence length $n$, num blocks $L$, steps per block $N$, confidence threshold $C$, exploration params $c^{\info}, \gamma, \alpha, \beta, N_e$.
\State $\bx_0 \leftarrow \left(\bx_{\text{prompt}}, \mask^{\otimes n}\right), ~t \leftarrow 0$ \Comment{Initialize sequence and step counter}
\For{$b = 1 \to L$}\Comment{Iterate over blocks}
    \State $t^b \leftarrow t$
    \While{$t - t^b \leq N$ and $\mathcal{M}_t \neq \emptyset$} \Comment{$N$ sampling budget per block}
        \State $\bx_{t+1} \leftarrow \textsc{Exploit}(\bx_t, b, C)$
        \Comment{\textbf{Exploitation}}
        \If{$\textsc{TriggerExploration}(\bx_{t+1}, b, \gamma, N_e)$} 
        \State $\bx_{t+1} \leftarrow \textsc{ImplicitExplore}(\bx_{t+1}, b-1)$
            \State $\bx_{t+2} \leftarrow \textsc{TargetedExplore}(\bx_{t+1}, b, \alpha, \beta, c^\info)$
        \Comment{\textbf{Targeted Exploration}}
            \State $t \leftarrow t+2$
        \Else
            \State $\bx_{t+1} \leftarrow \textsc{ImplicitExplore}(\bx_{t+1}, b)$
            \Comment{\textbf{Implicit Exploration}}
            \State $t \leftarrow t+1$
        \EndIf
    \EndWhile
    \If{$\mathcal{M}_t = \emptyset$} \textbf{break} \EndIf
\EndFor

\If{$\mathcal{M}_t \neq \emptyset$}\Comment{Final clean-up pass to exploit on remaining tokens}
    \For{$\_ = 1 \to n_f$}
    \State $\bx_{t+1} \leftarrow \textsc{Exploit}(\bx_t, L, C)$
        \Comment{\textbf{Clean-up Exploitation}}
    \State $t \leftarrow t+1$
    \EndFor
\EndIf
\State \textbf{Return} $\bx_t$
\end{algorithmic}
\end{algorithm}

\section{Experiments}\label{sec::experiment}

\subsection{Verification of Theorem \ref{thm:main}}\label{sec::experimentsetup}

\paragraph{Experimental Setup.}
We empirically verify the information-theoretic lower bound established in Theorem~\ref{thm:main} using the GSM8K mathematical reasoning dataset~\citep{cobbe2021gsm8k}. Our experiments employ \texttt{LLaDA-8B-Instruct}\footnote{\url{https://github.com/ML-GSAI/LLaDA}} as the base diffusion language model. We randomly sample $200$ test questions and configure the model with a maximum generation length of $512$ tokens and a block size of $64$ tokens per parallel decoding round.

For each sample, we execute standard confidence-based parallel decoding with five different factors: $f\in \{0.4,0.6,0.8,1.0,1.2\}$. To compute the true joint log-probability $-\log p(\bx)$ for each generated sequence, we perform autoregressive decoding (conditioning on all previous tokens) and accumulate the conditional log-probabilities token-by-token. We also evaluate the mathematical accuracy of each generated solution against the ground-truth answer, which yields similar average accuracy around $77\%$ across samples under different choices of $f$. \footnote{The accuracy is $78.5\%, 79.5\%, 78.5\%, 76\%, 76.5\%$ for $f=0.4,0.6,0.8,1.0,1.2$, respectively. }

\begin{figure}[h]
  \centering
  \begin{subfigure}{0.50\textwidth}
    \centering
    \includegraphics[width=\linewidth]{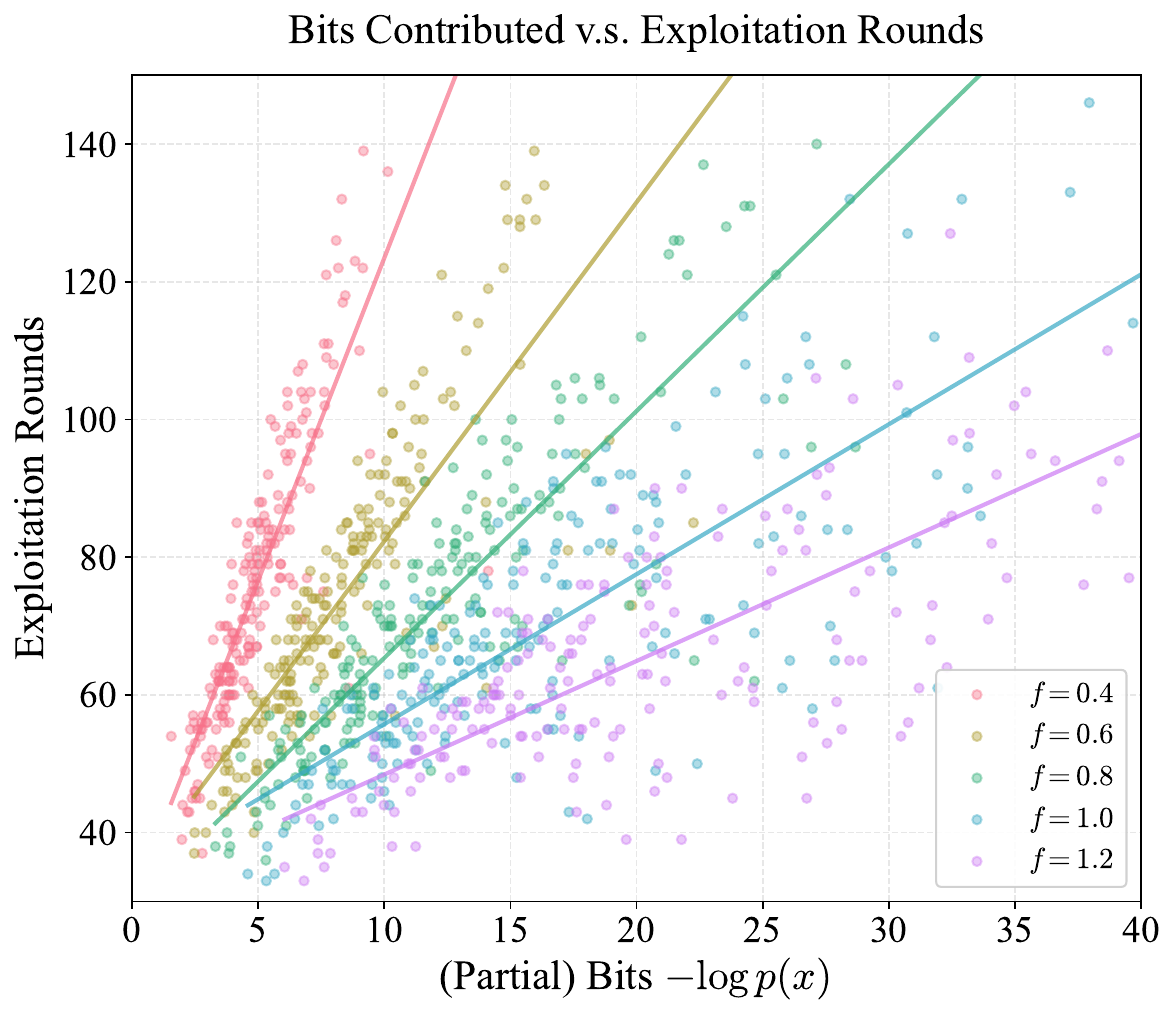}
    \caption{Exploitation rounds v.s. Bits per sample}
  \end{subfigure}\hfill
  \begin{subfigure}{0.50\textwidth}
    \centering
    \includegraphics[width=\linewidth]{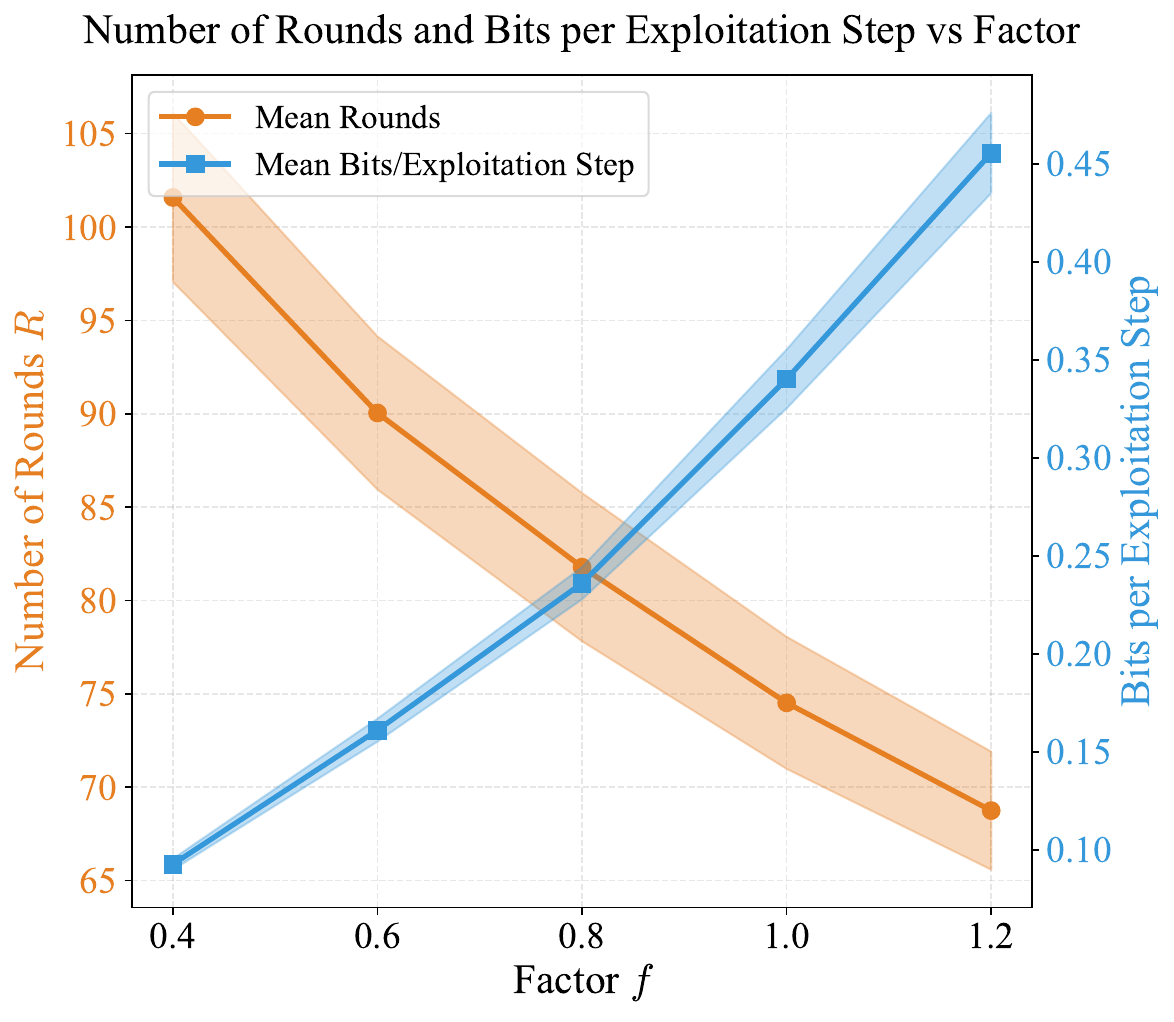}
    \caption{Exploitation rounds \& Bits per round}
  \end{subfigure}
  \caption{Panel (a): Exploitation rounds required versus partial log probability contributed. Each point represents a single sample from GSM8K dataset, with colors indicating different choice of factors $f$. We measure partial bits (excluding implicit exploration contributions) and exploitation rounds ($R - R_{\rm explore}$, excluding implicit exploration rounds) to isolate the effects of purely confidence-based parallel decoding. This decomposition removes the implicit exploration that occurs in confidence-based decoding, better demonstrating our theory; Panel (b): Number of rounds and bits per exploitation step as a function of factor $f$. Each point represents the mean over 200 fresh samples in GSM8K dataset for a given factor, with 95 \% confidence intervals shown as shaded regions. }
  \label{fig:rounds vs bits}
\end{figure}

\paragraph{Results and Analysis.}
Panel (a) in Figure~\ref{fig:rounds vs bits} presents the relationship between bits ($-\log p(\bx)$) and the rounds ($R$) across different confidence factors $f$. All five confidence factors exhibit clear linear relationships between bits and rounds, consistent with the lower bound in Equation~\eqref{eq:lowerbound}. The strong linear correlations validate that the number of decoding steps scales linearly with sequence information content. Moreover, the slopes of the fitted lines decreases monotonically with the confidence factor $f$, confirming that stricter confidence requirements (smaller $f$) lead to greater step inefficiency. The near-linear relationship between $f$ and the average bits per step also validates our theoretical per-round information budget.  Panel (b) further demonstrates the relation between the per step information decoded and confidence factor, demonstrating a nearly linear relationship between bits decoded per exploitation round and $f$, which substantiates our theory.

\subsection{Computation overhead of beam search}

\begin{figure}[h]
  \centering
  \begin{subfigure}{0.48\textwidth}
    \centering
    \includegraphics[width=\linewidth]{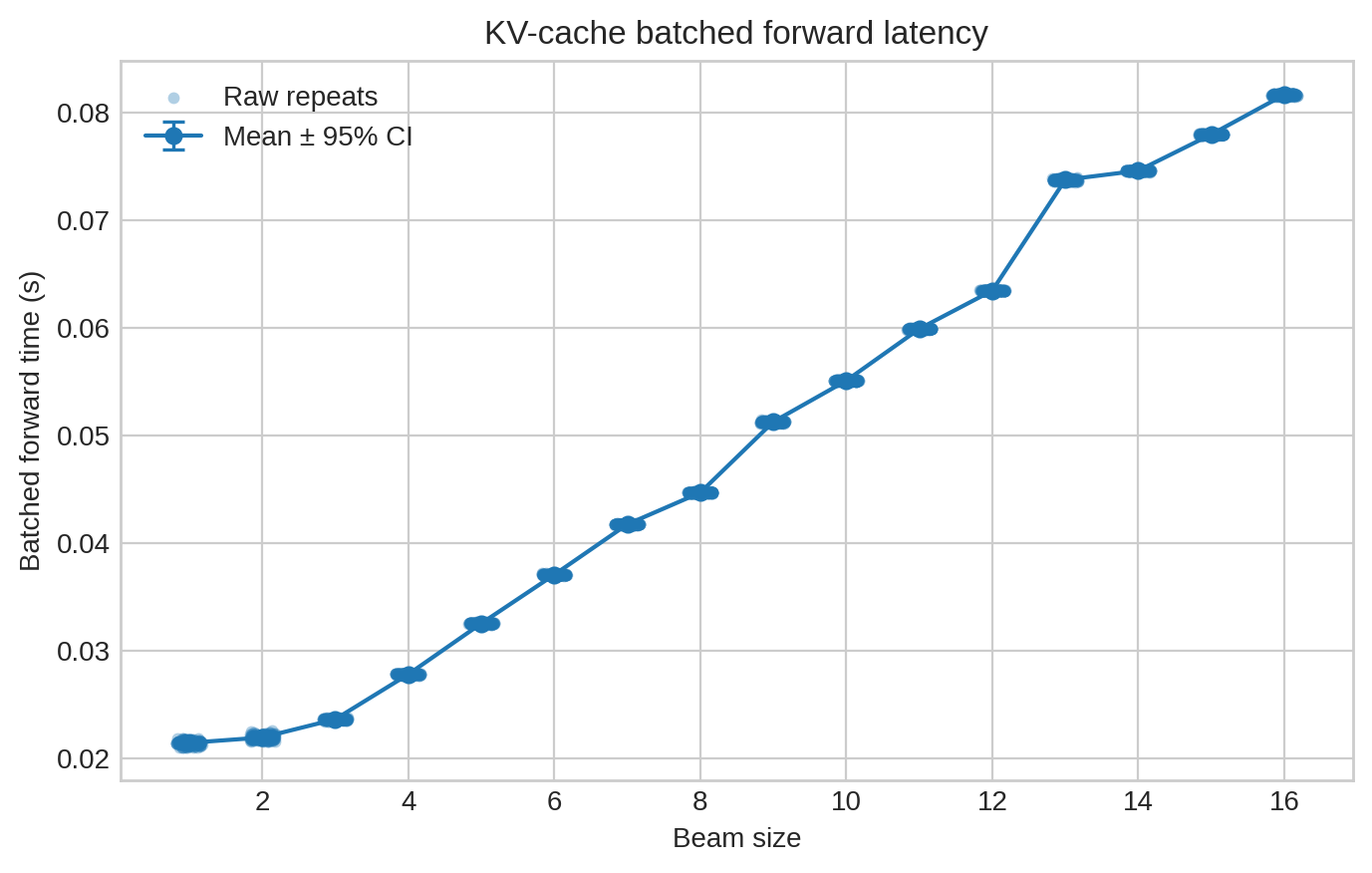}
    \caption{Average wallclock on H100}
  \end{subfigure}\hfill
  \begin{subfigure}{0.48\textwidth}
    \centering
    \includegraphics[width=\linewidth]{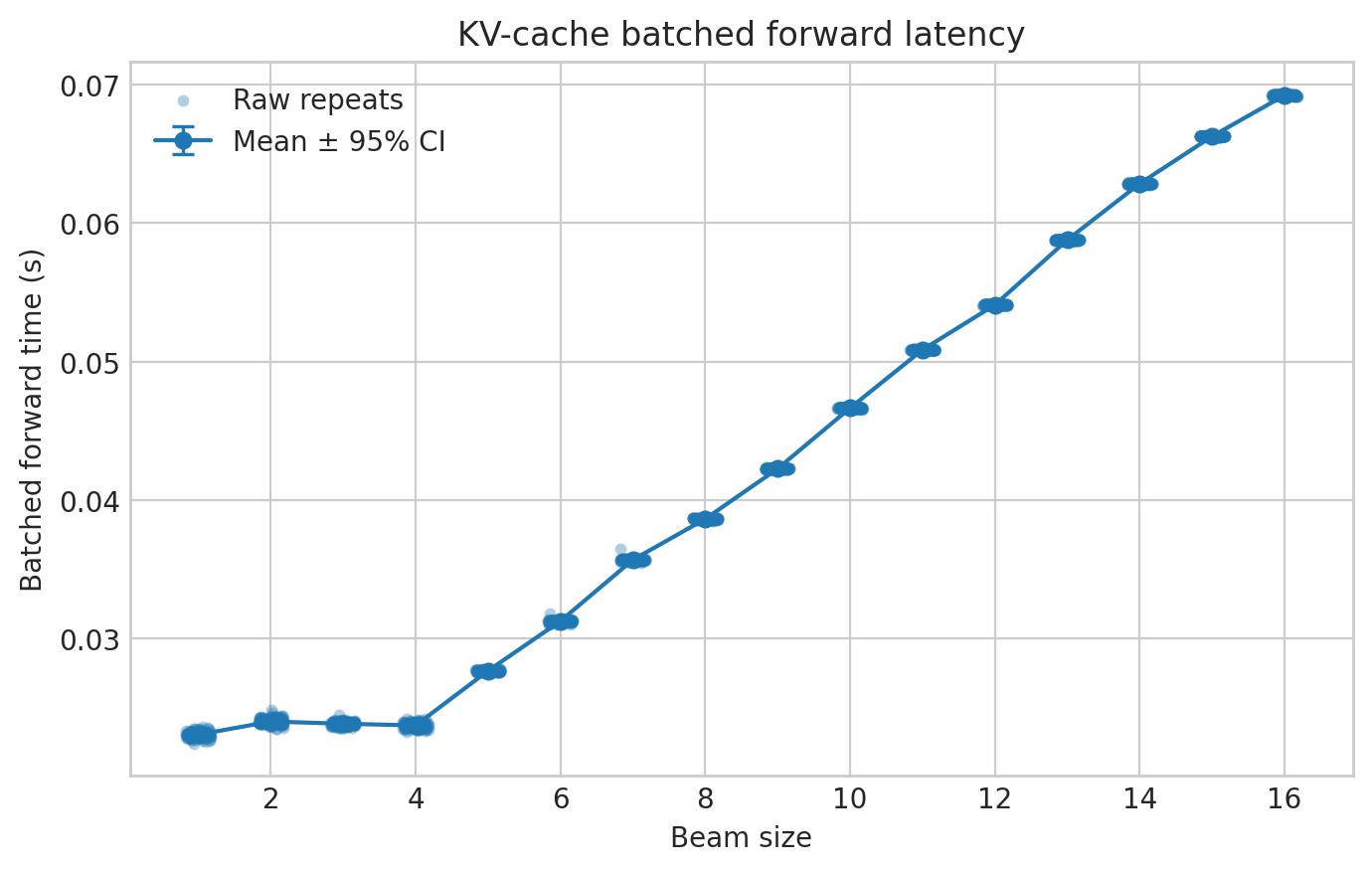}
    \caption{Average wallclock on B200}
  \end{subfigure}
  \caption{Average wall-clock time for batched forward passes with different beam sizes under KV cache on NVIDIA H100 and B200 GPU. We adapted the Fast-DLLM codebase  for KV-cache implementation. To simulate realistic decoding scenarios, we generate sequences of length 512 with block length 64 (8 blocks total), using Fast-DLLM's confidence-based decoding algorithm. At the first 4 decoding steps of each block and for each beam size $k$, we perform an exploration round by selecting $k$ candidate tokens and creating $k$ hypothesis sequences through position substitution in the current block. We then perform a batched forward pass on the $k$ hypothesis sequences using the shared KV cache, repeating each batched forward 10 times to compute the average latency. This yields 32 exploration rounds in total (4 steps $\times$ 8 blocks), from which we compute the mean and 95\% confidence interval across different exploration positions for each beam size.}
  \label{fig:compute-cache-batch}
\end{figure}

We further investigate the computational overhead introduced by beam search under varying batch sizes. Figure~\ref{fig:compute-cache-batch} reports the average batched forward latency with KV cache for the LLaDA-8B-Instruct model as a function of beam size on an H100 GPU (we use the Fast-DLLM\footnote{\url{https://github.com/NVlabs/Fast-dLLM}} implementation for KV caching). We observe a substantial ``free lunch" region in which increasing the batch size incurs negligible additional cost: approximately 2 on H100 and 4 on B200. Within this regime, the computational cost of exploring multiple candidate tokens in a single batched forward pass is comparable to that of a single sequential forward pass. This ``free lunch" region arises because memory bandwidth, rather than computation, becomes the bottleneck at small batch sizes. Even beyond this regime, the marginal cost per additional beam remains relative small (approximately 0.005 seconds), which is significantly less than the cost of a single non-batched forward pass (over 0.02 seconds).

Based on this observation, we ensure that the hyperparameters for ETE we report remain below this ``free lunch" region of a batch size of 4, to ensure that the wall-clock cost of each step (batched or otherwise) remains the same.

\subsection{Selection Of Efficiency Metric}

The primary objective of this work is to isolate and quantify the algorithmic efficiency of the Explore-Then-Exploit (ETE) framework relative to standard confidence-based decoding. To this end, we adopt the \textbf{number of model forward passes} (often denoted as NFE) as our principal metric for efficiency. We showcase the efficiency of our method using the reduction in the number of NFEs.

While metrics such as User Tokens Per Second (User-TPS) or end-to-end wall-clock time are frequently used in deployment contexts, they are suboptimal for measuring fundamental algorithmic improvements. These time-based metrics inevitably conflate the approach's theoretical efficiency with extrinsic variables, most notably implementation maturity. For example, the overhead of unoptimized Python code relative to established baselines obscures the intrinsic gains of the novel sampling strategies.

Furthermore, we constrain the hyperparameters of our ETE approach to lie strictly within the ``free lunch'' computational regime (typically batch sizes $\le 4$ on modern hardware). In this regime, the execution cost of a batched forward pass is effectively equivalent to that of a single sequence pass. Consequently, the total count of forward passes serves as a robust and faithful proxy for real-world performance, providing a direct measure of decoding efficiency that is invariant to the current optimization level of the experimental codebase.

\subsection{Benchmark results}

To comprehensively assess the effectiveness of our approach, we evaluate it on four widely used benchmarks: MMLU~\citep{wang2024mmlu}, GSM8K~\citep{cobbe2021gsm8k}, HumanEval~\citep{chen2021evaluatinglargelanguagemodels}, and MATH~\citep{hendrycksmath2021}. We adopt Confidence-Aware Parallel Decoding~\citep{wu2025fast} with a static threshold strategy as our baseline, as we found this configuration yields the strongest accuracy-speed frontiers among prior methods.

All experiments use LLaDA-8B-Instruct\footnote{\url{https://huggingface.co/GSAI-ML/LLaDA-8B-Instruct}} as the sampling model with a block size of $64$. We set the generation length to $512$ for MATH, GSM8K, and HumanEval, and $256$ for MMLU-Pro. For MMLU-Pro, we evaluate on a uniform subset of 1000 samples to align the dataset size with other benchmarks and ensure computational feasibility.

\begin{figure}[h]
  \centering
  \begin{subfigure}{0.5\textwidth}
    \centering
    \includegraphics[width=\linewidth]{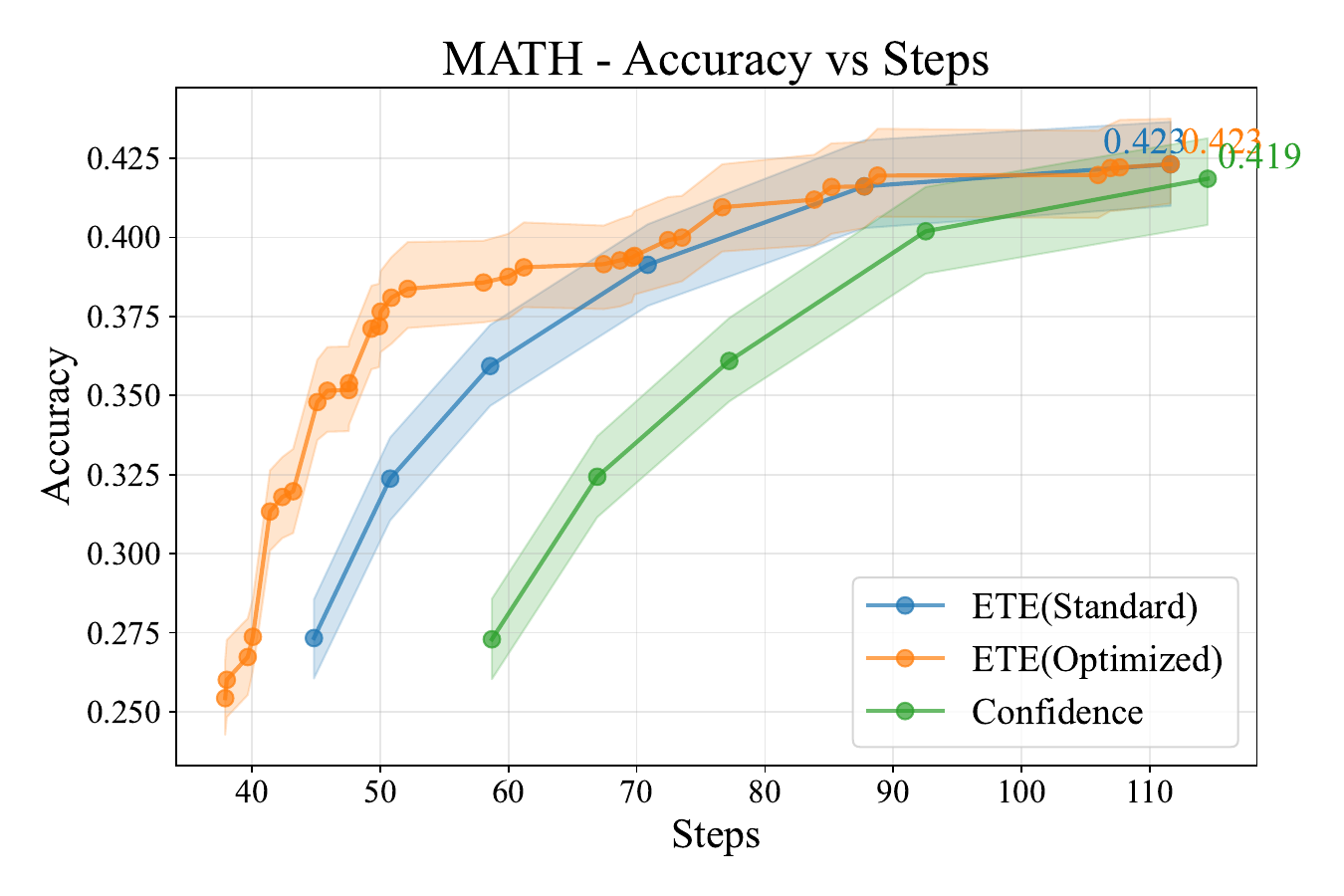}
    \caption{MATH}
  \end{subfigure}\hfill
  \begin{subfigure}{0.5\textwidth}
    \centering
    \includegraphics[width=\linewidth]{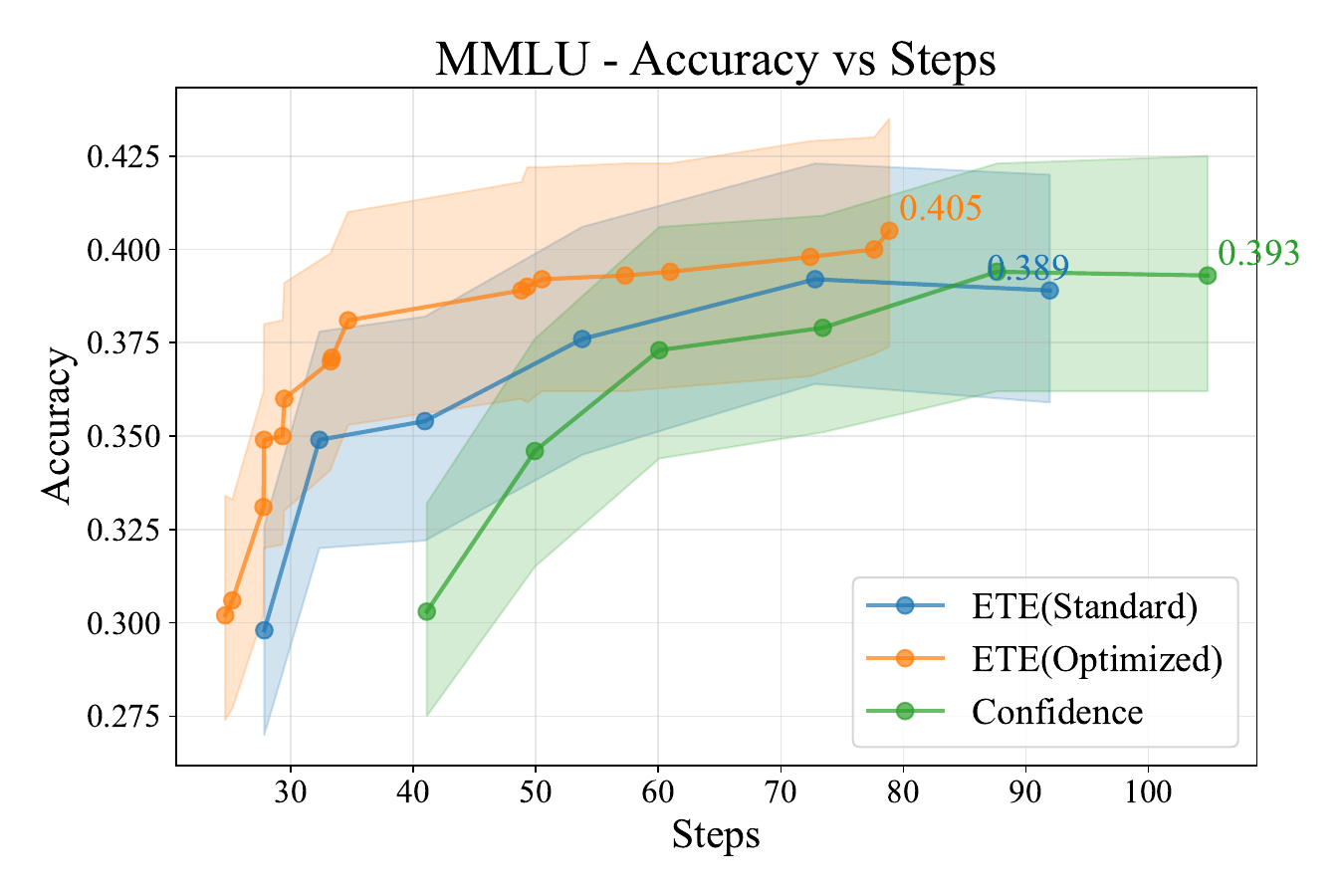}
    \caption{MMLU-Pro}
  \end{subfigure}
  \begin{subfigure}{0.5\textwidth}
    \centering
    \includegraphics[width=\linewidth]{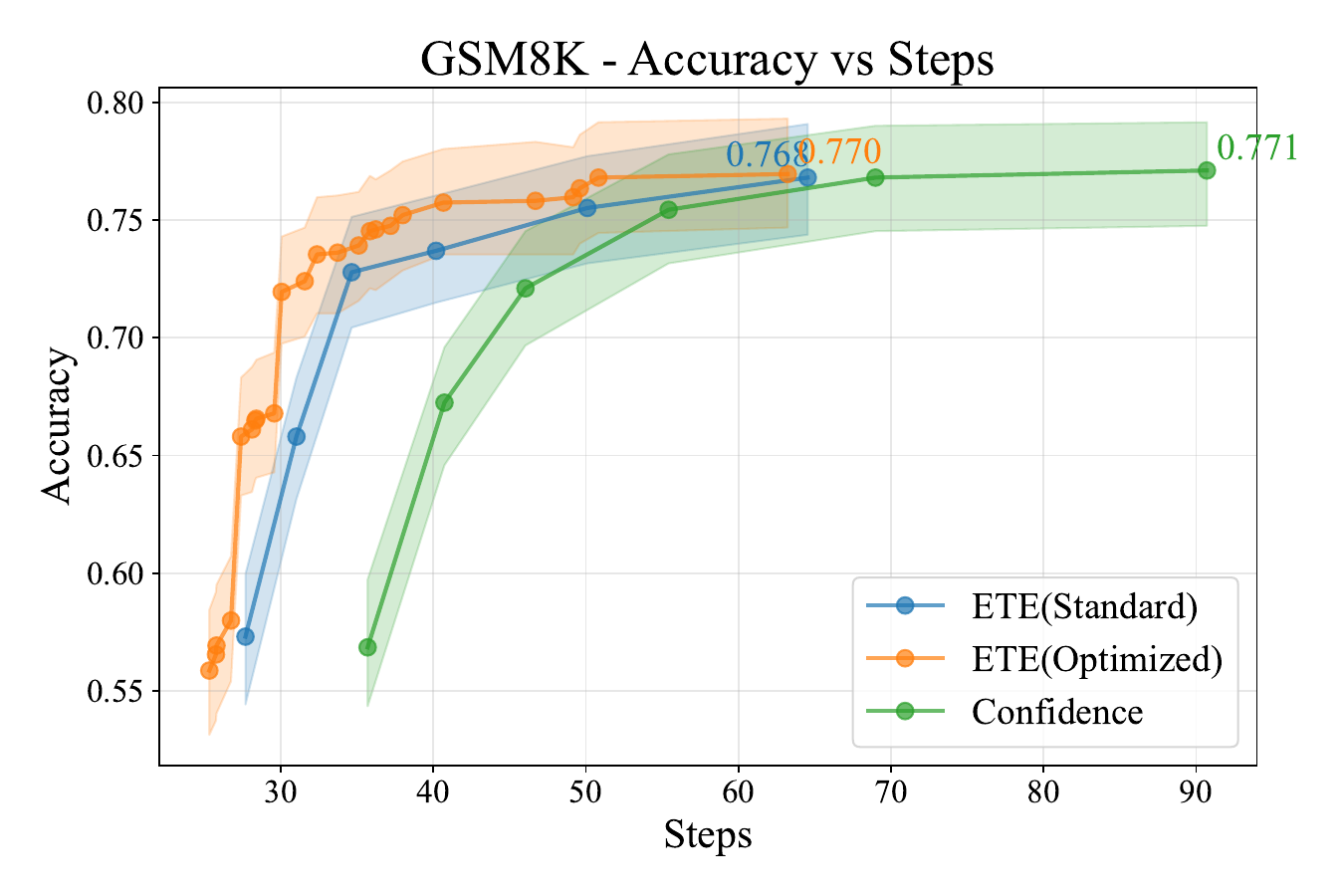}
    \caption{GSM8K}
  \end{subfigure}\hfill
  \begin{subfigure}{0.5\textwidth}
    \centering
    \includegraphics[width=\linewidth]{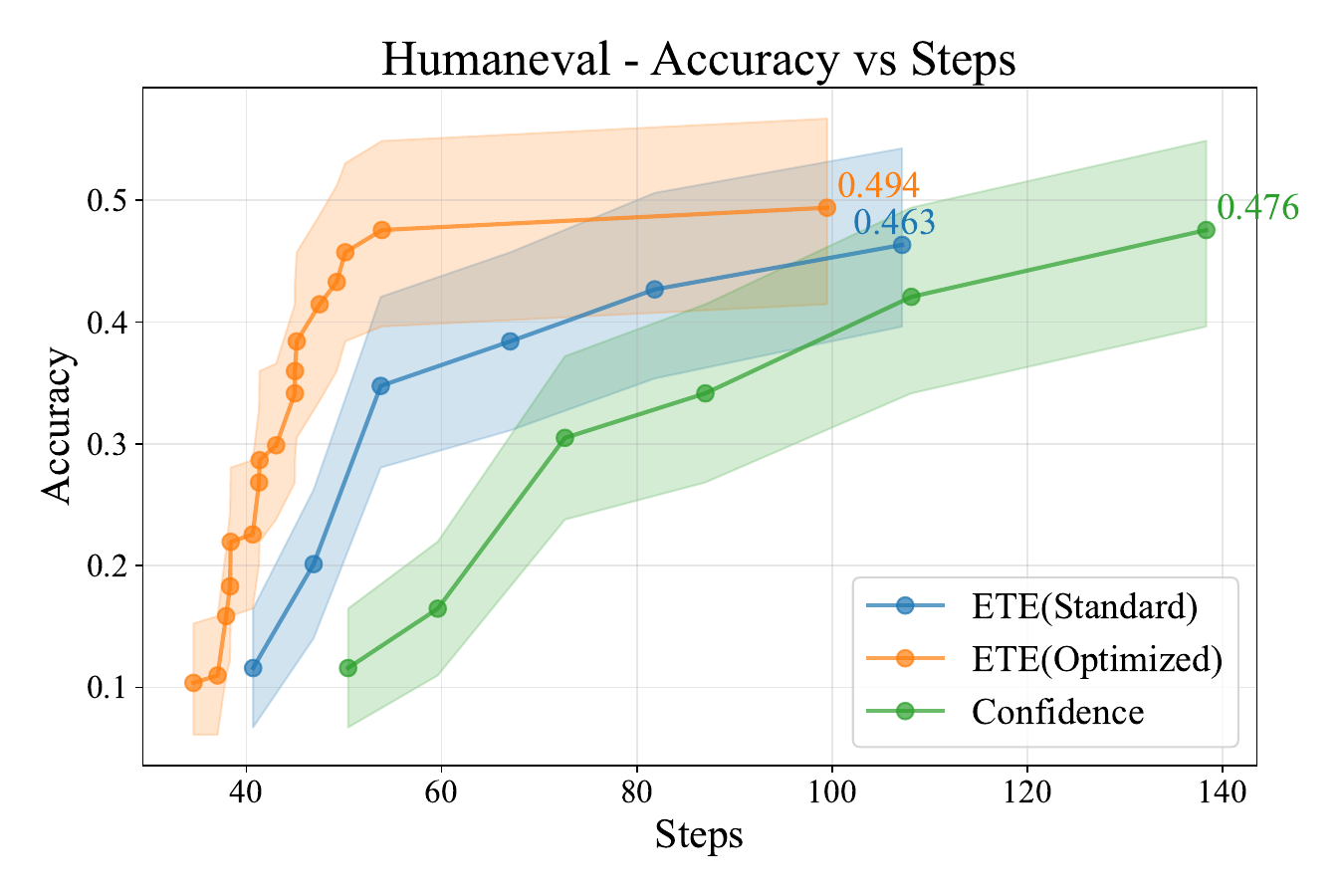}
    \caption{HumanEval}
  \end{subfigure}
  \caption{Accuracy-steps frontiers of our method versus baseline on four benchmarks.}
  \label{fig:frontiers}
\end{figure}

To analyze the tradeoff between performance and computational cost, we compare the methods across a range of confidence thresholds and step budgets. For the baseline, we report the Pareto frontier of test metrics across varying confidence thresholds. For \name{}, we report two performance curves:
\begin{itemize}
    \item \textbf{Controlled Comparison (`ETE (Standard)'):} We measure performance across a range of confidence thresholds while holding other hyperparameters fixed (tuned on the GSM8K training set). This ensures a direct and fair comparison with the baseline by varying the same primary hyperparameter (the static confidence threshold).
    \item \textbf{Optimal Frontier (`ETE (Optimized)'):} We report the Pareto frontier formed by varying both confidence thresholds and step budgets. This curve illustrates the full potential of \name{} by leveraging the additional degree of freedom provided by the fast block diffusion sampling scheme described in Section~\ref{sec:fast-block-diffusion}.
\end{itemize}

\paragraph{Tradeoff between accuracy and total inference steps.}

Figure~\ref{fig:frontiers} illustrates the tradeoff between generation accuracy and the total number of inference steps. Shaded regions denote 95\% confidence intervals computed via bootstrapping. Across all four benchmarks, we observe that the \name{} frontier is consistently superior to that of the confidence-based baseline. In direct comparisons using fixed hyperparameters, the \name{} (Standard) attains higher average accuracy at comparable or lower step budgets, highlighting statistically significant improvements particularly in low-compute regimes. Moreover, by leveraging the additional degree of freedom in step scheduling, the \name{} (Optimized) achieves an even more favorable tradeoff, reaching higher peak accuracies with fewer steps. Overall, these results demonstrate that \name{} allocates computational resources more effectively, translating to superior accuracy-per-step efficiency across diverse domains.

\section{Conclusions and Future Works}

This work establishes both theoretical and empirical foundations for improving parallel decoding in diffusion language models. We have proven a fundamental lower bound showing the inherent inefficiency of confidence-based parallel decoding algorithm  from the information theoretical viewpoint. Motivated by this insight, we have developed \name{}, a training-free algorithm that explicitly targets high-entropy tokens through fast block diffusion sampling and principled exploration mechanisms. Experiments across four benchmarks have demonstrated that \name{} consistently outperforms confidence-based baselines in both decoding efficiency and output quality.

Regarding future works, we can pursue several promising directions in both theory and practice. Theoretically, establishing upper bounds on required rounds and computational overheads under structured data assumptions remains a promising question, which naturally connects to the broader problem of parallel sequential sampling with limited  queries on marginal distributions \citep{anari2024parallel,hu2025diffusionmodelssecretlyexchangeable}. Algorithmically, learning-based  exploration strategies--such as training selection or scoring heads--could replace look-ahead computation while maintaining or improving quality.

\clearpage
\bibliography{ref, dllm_ref}
\ifdefined\isarxiv
\bibliographystyle{iclr2024_conference}
\else
\bibliographystyle{iclr2024_conference}
\fi

\newpage
\appendix

\onecolumn

\section{Proof of \Cref{thm:main}}
\label{sec:proof}

\begin{proof}
We first control the approximation error incurred by factorizing the joint likelihood into per-group conditionals. For any ordered partition $(A_1,\ldots,A_R)$ as above, by the ordered-partition chain rule of probability density function, we have
\begin{align*}
-\log p(\bx) \;=\; \sum_{r=1}^{R} \bigl[-\log p(x^{A_r}\mid x^{C_r})\bigr] .
\end{align*}
 For each round $r\in[R]$, by Assumption~\ref{asp:dynamic_threshold} and Fréchet inequality, the joint conditional probability of $x^{A_r}$ given the past unmasked set $C_r$ can be lower bounded by 
 \begin{align}
   p(x^{A_r} \mid x^{C_r}) \ge 1- \sum_{i \in A_r} (1-p(x^i|x^{C_r}))\ge 1-\frac{f|A_r|}{1+|A_r|} . \label{equ::range of joint likelihood}
 \end{align}
Thus, the entire negative joint log probability $-\log p(\bx)$ can be bounded by 
\begin{align*}
       -\log p(\bx)&=\sum_{r=1}^{R} \bigl[-\log p(x^{A_r}\mid x^{C_r})\bigr] \\
       &\leq \sum_{r=1}^{R} \log \left(\frac{1}{1-f|A_r|/(1+|A_r|)}\right)\tag{plugging in \eqref{equ::range of joint likelihood}}\\
       &=\sum_{r=1}^{R} \log \left(\frac{1+|A_r|}{1+(1-f)|A_r|}\right) \\
       &\le R \log \left(\frac{1+n}{1+(1-f)n}\right) \tag{$\frac{1+t}{1+(1-f)t}$ is monotone increasing with $t>0$}.
\end{align*}
Rearranging, we obtain
\begin{align}\label{eq:bits-bound-1}
    R \geq \frac{-\log p(\bx)}{\log \left(\frac{1+n}{1+(1-f)n}\right)}.
\end{align}
On the other hand, by the definition of $\epsilon$ in \eqref{eq:epsilon}, we have
\begin{align}
        \left|\sum_{r=1}^{R} \log p(x^{A_r}\mid x^{C_r})- \sum_{r=1}^{R} \sum_{i\in A_r} \log p(x^i\mid x^{C_r})\right| \le \epsilon  \label{equ::eps recall}
\end{align}
for any $r\in[R]$.
Consequently, again by applying the chain rule and summing up over $r$, we have
\begin{align*}
     -\log p(\bx)&=\sum_{r=1}^{R} \left[-\log p(x^{A_r}\mid x^{C_r})-\sum_{i\in A_r} \bigl[-\log p(x^i\mid x^{C_r})\bigr]+\sum_{i\in A_r} \bigl[-\log p(x^i\mid x^{C_r})\bigr]\right]\\
     &\le \epsilon+ \sum_{r=1}^{R} \sum_{i\in A_r} \bigl[-\log p(x^i\mid x^{C_r})\bigr] \tag{plugging in \eqref{equ::eps recall}}\\
     &\le \epsilon+ \sum_{r=1}^{R} \left[-\sum_{i\in  A_r}\log\left(1-\frac{f}{1+|A_r|}\right)\right] \tag{by Assumption \ref{asp:dynamic_threshold}}\\
     &= \epsilon+ \sum_{r=1}^{R} \left[|A_r|\log \left(1+\frac{f}{(1-f)+|A_r|}\right)\right]\\
     &\le \epsilon+ \sum_{r=1}^{R} \frac{|A_r|f}{1-f+|A_r|}
     \tag{$\log (1+t) \le t$ for $t\ge 0$.}\\
&\le \epsilon+ R f.
\end{align*}
Rearranging, we obtain
\begin{align}\label{eq:bits-bound-2}
    R \geq \frac{-\log p(\bx) - \epsilon}{f}.
\end{align}
Combining Eq.~\eqref{eq:bits-bound-1} and Eq.~\eqref{eq:bits-bound-2}, we have
\begin{align*}
    R \geq \max \left\{\frac{-\log p(\bx)}{\log \left(\frac{1+n}{1+(1-f)n}\right)}, \frac{-\log p(\bx) - \epsilon}{f}\right\}.
\end{align*}
This completes the proof.
\end{proof}

\clearpage

\section{Detailed Subroutines in Algorithm \ref{alg:main}}

\begin{algorithm}[h]
\caption{\textsc{\name{}} Exploitation and Exploration Subroutines}
\label{alg:subroutines}
\begin{algorithmic}[1]

\Procedure{Exploit}{$\bx_t, b, C$}
    \State Form the feasible set $S_t$ as in Eq.~\eqref{eq:feasible-pos}.
    \State Compute exploitation tokens $\bx_{t+1}$ according to Eq.~\eqref{eq:exploit-conf}.
    \State \Return $\bx_{t+1}$
\EndProcedure

\vspace{0.5em}

\Procedure{ImplicitExplore}{$\bx_{t+1}, b$}
    \State Reuse $c(\hat x_{t+1}^i)$ from the last forward pass.
    \For{$b' = 1,\dots,b$} \Comment{Parallel implicit exploration across active blocks}
        \State $S_{b';t} \gets S_t \cap \mathcal{B}_{b'}$
        \If{$S_{b';t} = \emptyset$} \textbf{continue} \EndIf
        \State Let $U_{b';t}$ be the subset of $S_{b';t}$ not already committed by Eq.~\eqref{eq:exploit-conf}.
        \If{$U_{b';t} \neq \emptyset$}
            \State $i^\star \gets \arg\max_{i \in U_{b';t}} p^{i}_\theta(\hat x_{t+1}^i \mid \bx_t)$
            \State $x_{t+1}^{i^\star} \gets \hat x_{t+1}^{i^\star}$ \Comment{One highest-confidence token per block}
        \EndIf
    \EndFor
    \State \Return $\bx_{t+1}$
\EndProcedure

\vspace{0.5em}

\Procedure{TriggerExplore}{$\bx_{t+1}, b, \gamma, N_e$}
\State $i_{\rm frontier}=\min(bn_b,\max((b-1)n_b,\arg\max_{i}\{i:x_{t}^i \neq \mask\})+n_b/2)$
\State$\mathcal{W}_t\gets\{i:x_t^i=\mask, i \in ((b-1)n_b,i_{\rm frontier}]\}$ . \Comment{Identify the frontier window}
    \State $\bar c_t \gets \frac{1}{|\mathcal{W}_t|} \sum_{i \in \mathcal{W}_t} p^{i}_\theta(\hat x_{t+1}^i \mid \bx_t)$. \Comment{Compute the average frontier confidence}
    \State Let $R_{b;t}$ be the number of masked tokens remaining in block $b$.
    \If{$\bar c_t < \gamma$ \textbf{and} $R_{b;t} > N_e$ \textbf{and} exploration budgets are not exhausted}
        \State \Return \textbf{True} \Comment{Frontier is information-poor and far from completion}
    \Else
        \State \Return \textbf{False}
    \EndIf
\EndProcedure

\vspace{0.5em}

\Procedure{TargetedExplore}{$\bx_{t+1}, b, \alpha, \beta, C_{\info}$}
    \State Construct $\mathcal{H}_t$ using Eq.~\eqref{eq:explore-space}.
    \For{$j \in \mathcal{H}_t$}\Comment{Form search beam}
    \State $\bx_{t+1;j} \leftarrow \bx_{t+1}, ~x_{t+1;j}^j=\arg\max_{v \in \V}p_\theta^j(v \mid \bx_t)$.
    \EndFor
    \State Compute $(\bx_{t+2;j})_{j \in \mathcal{H}_t} \leftarrow \textsc{Exploit}((\bx_{t+1;j})_{j \in \mathcal{H}_t}, b,C)$ in batch.
    \State Select $j^\star \gets \arg\max_{j} s(j)$ for the score in Eq.~\eqref{eq:explore-score}.
    \State \Return $\bx_{t+2;j^*}$
\EndProcedure

\end{algorithmic}
\end{algorithm}

\end{document}